\documentclass[10pt,journal]{IEEEtran}

\usepackage{afterpage}
\usepackage{algorithmicx}
\usepackage[linesnumbered,ruled,noend]{algorithm2e}
\SetKwInOut{Input}{Input}
\SetKwInOut{Output}{Output}

\SetKwProg{Fn}{Function}{\string:}{end}

\usepackage[backend=biber, natbib=true, style=ieee, url=false, doi=false, sorting=none]{biblatex}

\usepackage[a4paper, margin=1.82cm]{geometry}

\usepackage{lipsum}
\usepackage{amsfonts}
\usepackage{amsmath}
\usepackage{amssymb}
\usepackage{array}
\usepackage{bm}
\usepackage{booktabs}
\usepackage{boxedminipage}
\usepackage{colortbl}
\usepackage{enumerate}
\usepackage{extarrows}
\usepackage{float}
\usepackage{graphics}
\usepackage{graphicx}
\usepackage{indentfirst}
\usepackage{latexsym}
\usepackage{makecell}
\usepackage{mismath}
\usepackage{multirow}
\usepackage{picinpar}
\usepackage{placeins}
\usepackage{subfigure}
\usepackage{textcomp}
\usepackage{theorem}
\usepackage[para]{threeparttable}
\usepackage[normalem]{ulem}
\usepackage{url}
\usepackage{wrapfig}
\usepackage{xcolor}

\newtheorem{example}{Example}

\newcommand{\qed}{\hspace*{\fill}$\Box$}

\IEEEoverridecommandlockouts

\allowdisplaybreaks

\newcommand{\tit}[1]{{\textit{#1}}}
\newcommand{\tbf}[1]{{\textbf{#1}}}
\newcommand{\ie}{\textit{i.e.}}
\newcommand{\eg}{\textit{e.g.}}
\newcommand{\wrt}{\textit{w.r.t.}}

\newcommand{\etal}{\textit{et al.}}

\newcommand{\cadd}[1]{{\color{black}{#1}}}

\begin{document}

\title{\cadd{Gradient Estimation Methods} of Approximate Multipliers for High-Accuracy Retraining of Deep Learning Models}
\author{Chang~Meng,~\IEEEmembership{Member, IEEE},
Wayne~Burleson,~\IEEEmembership{Fellow, IEEE}, 
and Giovanni~De~Micheli,~\IEEEmembership{Life~Fellow, IEEE}
\thanks{Chang~Meng is with the Department of Mathematics and Computer Science, Eindhoven University of Technology, The Netherlands (email: c.meng@tue.nl).}
\thanks{Wayne~Burleson is with the Department of Electrical and Computer Engineering, University of Massachusetts Amherst, USA (email: burleson@umass.edu).}
\thanks{Giovanni~De~Micheli is with the Integrated Systems Laboratory, École Polytechnique Fédérale de Lausanne, Switzerland (email: giovanni.demicheli@epfl.ch).}
\thanks{Corresponding author: Chang Meng.}}

\maketitle

\begin{abstract}

Approximate multipliers (AppMults) are widely used in deep learning accelerators to reduce their area, delay, and power consumption.
However, AppMults introduce arithmetic errors into deep learning models,
necessitating a retraining process to recover accuracy.
A key step in retraining is computing the gradient of the AppMult,
\ie{}, the partial derivative of the approximate product with respect to each input operand.
Existing approaches typically estimate this gradient using that of the accurate multiplier (AccMult),
which can lead to suboptimal retraining results.
To address this, we propose two methods to obtain more precise gradients of AppMults.
The first, called \tit{LUT-2D}, characterizes the AppMult gradient with 2-dimensional lookup tables (LUTs),
providing fine-grained estimation and achieving the highest retraining accuracy.
The second, called \tit{LUT-1D},
is a compact and more efficient variant that
stores gradient values in 1-dimensional LUTs,
achieving comparable retraining accuracy with shorter runtime.
Experimental results show that
on CIFAR-10 with convolutional neural networks,
\cadd{our LUT-2D and LUT-1D methods improve retraining accuracy by 3.83\% and 3.72\% on average, respectively.}
On ImageNet with vision transformer models,
\cadd{our LUT-1D method} improves retraining accuracy by 23.69\% on average,
compared to a state-of-the-art retraining framework.

\end{abstract}


\section{Introduction}\label{sect:intr}

Modern \tit{artificial intelligence} (\tit{AI}) technologies excel in a wide range of areas such as natural language processing and computer vision. 
This success drives widespread adoption of deep learning accelerators,
such as \tit{convolutional neural network (CNN)} accelerators and transformer accelerators, in edge devices and cloud systems.
However, this rapid growth raises serious concerns about power consumption~\cite{schwartz2020green}.

To achieve energy-efficient deep learning accelerators,
researchers have adopted an emerging design paradigm called \tit{approximate computing},
which reduces power consumption at the cost of errors~\cite{han2013approximate,mittal2016survey}.
Approximate computing is particularly suitable for deep learning accelerators,
since they are inherently resilient to errors and noise.
By carefully introducing errors into the hardware,
the final output quality can remain nearly unaffected,
while area, delay, and power consumption are significantly reduced~\cite{armeniakos2022hardware}.
Generally speaking,
traditional model compression techniques with low-precision data representations,
such as int4~\cite{choukroun2019low}, float8~\cite{park2021neural} and float4~\cite{dettmers2023qlora},
can be viewed as approximate computing techniques.

\begin{figure}[!htbp]
    \centering
    \includegraphics[width=1.00\columnwidth]{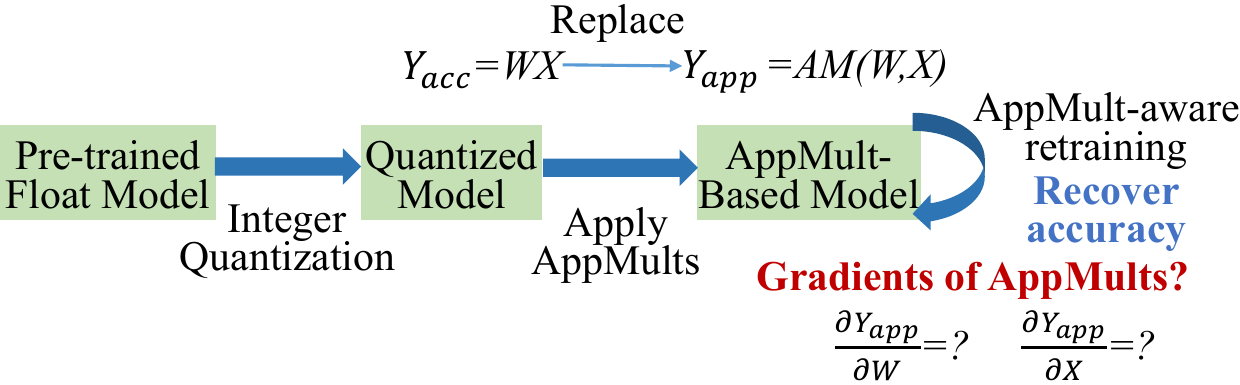}
    \caption{Design flow of AppMult-based deep learning accelerators.}
    \label{fig:motivation}
\end{figure}

Among various approximation techniques,
those based on \tit{\cadd{approximate multipliers (AppMults)}} are among the most popular~\cite{armeniakos2022hardware,jiang2020approximate}.
Fig.~\ref{fig:motivation} shows a typical design flow for AppMult-based deep learning accelerators.
It begins with a pre-trained floating-point deep learning model, 
followed by quantization to convert both model weights and activations into integers.
Since integer multipliers are \cadd{the most} power-consuming in a quantized accelerator,
they are replaced with approximate versions to reduce power consumption.
However, the inaccuracies of AppMults degrade the model accuracy,
necessitating AppMult-aware retraining to recover accuracy.

A fundamental question in AppMult-aware retraining is how to compute the gradient of an AppMult,
\ie{}, the partial derivative of the approximate product $Y_\tit{app}$
\tit{with respect to (\wrt{})} each input operand, $W$ or $X$.
Prior works adopt the \tit{straight-through estimator (STE)},
which approximates the gradient of an AppMult using that of an \tit{accurate multiplier (AccMult)}~\cite{he2018axtrain,de2020proxsim,qian2021approximate,danopoulos2022adapt,gong2023approxtrain,novkin2023approximation,yu2024toward,danopoulos2025transaxx}.
This approach is effective for small-error AppMults,
whose gradients closely match those of AccMults.
However, in scenarios requiring aggressive power reduction,
AppMults with relatively large errors are preferred.
For such designs, their gradients can deviate significantly from those of AccMults,
causing
the STE-based retraining to yield suboptimal results.
To address this issue, we propose more precise gradient estimation methods for AppMults,
aiming to improve retraining accuracy,
especially for AppMults with relatively large errors.

Our main contributions are as follows:

\begin{itemize}
    \item We develop an AppMult-aware retraining framework supporting user-defined gradients,
    applicable to CNNs and transformers.
    \item We propose two techniques to estimate more precise gradients of AppMults.
    The first, \tit{LUT-2D}, is fine-grained and achieves the best retraining accuracy.
    The second, \tit{LUT-1D}, is more efficient while still achieving high retraining accuracy.
    \item \cadd{Our method improves the retraining accuracy across various models and datasets.
    On CIFAR-10 with CNNs,
    our LUT-2D and LUT-1D methods improve accuracy by 3.83\% and 3.72\% on average, respectively.
    On ImageNet with vision transformers,
    our LUT-1D method improves accuracy by 23.69\% on average.}
\end{itemize}

Our AppMult-aware retraining framework is available at\\{\footnotesize\url{https://github.com/changmg/AppMult-Aware-Retraining/tree/journal}}

Compared to our preliminary version~\cite{meng2025gradient},
this paper introduces two major extensions.
First, we propose the LUT-1D method,
which accelerates retraining and reduces memory requirements.
Second, we provide more comprehensive experiments,
including new evaluations on vision transformers and low-bit quantization schemes with the ImageNet dataset.

The remainder of the paper is organized as follows.
Section~\ref{sect:related-works} reviews related works.
Section~\ref{sect:prel} introduces preliminaries of AppMult-aware retraining.
Sections~\ref{sect:framework} and~\ref{sect:grad} present our AppMult-aware retraining framework and the proposed gradient estimation methods, respectively.
Section~\ref{sect:result} reports experimental results.
Section~\ref{sect:concl} concludes the paper.
\section{Related Works}\label{sect:related-works}

The use of AppMults in deep learning accelerators has been studied for more than a decade.
This section reviews prior works from three aspects:
1) AppMult designs for deep learning accelerators,
2) AppMult-aware retraining frameworks, and 
3) methods that assign different AppMults to different network layers.
Our work focuses on the second aspect:
a general retraining framework that improves the accuracy of models using existing AppMults.

\subsection{AppMult Designs for Deep Learning Accelerators}
Many studies design AppMults tailored to deep learning workloads.
For a broader survey of AppMults, see~\cite{wu2024survey}.
Below we discuss design strategies, targeted models, and datasets used in each representative work.

Mrazek~\etal{}~\cite{mrazek2016design} applied Cartesian Genetic Programming to create power-efficient AppMults for \tit{neural networks (NNs)},
evaluating LeNet-6~\cite{lecun1998gradient} and a small \tit{multilayer perceptron (MLP)} using the MNIST~\cite{lecun1998mnist} and SVHN~\cite{netzer2011reading} datasets.
They later released \tit{EvoApproxLib}~\cite{mrazek2017evoapprox8b}, 
a library including signed/unsigned AppMults with multiple bit widths and approximation levels.
Ahmadinejad~\etal{}~\cite{ahmadinejad2021energy} proposed energy- and quality-efficient AppMults
using NAND-based complemented partial products and new compressors.
They tested their AppMults on LeNet-5 and a small MLP with MNIST and SVHN.
Kim~\etal{}~\cite{kim2018efficient} built energy-efficient AppMults based on Mitchell's log multiplication,
optimized for CNN inference.
They evaluated on MNIST, CIFAR-10~\cite{krizhevsky2009learning}, and ImageNet~\cite{deng2009imagenet}.
Kumari~\etal{}~\cite{kumari2025design} introduced 8-bit AppMults via recursive, bit-wise, and hybrid approximation using partial bit OR.
They provided 15 accuracy-hardware cost tradeoff points
and tested the AppMults on an MLP with MNIST.
Hu~\etal{}~\cite{hu2024configurable} analyzed correlations between \tit{partial products (PPs)} in multipliers
and applied PP speculation to design accuracy-configurable AppMults.
They evaluated CNNs such as VGG-16, ResNet-18 and ResNet-50 on CIFAR-10 and ImageNet.
For NN inference with AppMults, \tit{TFApprox}~\cite{vaverka2020tfapprox} provides a LUT-based simulator for 8-bit AppMults on GPUs using TensorFlow~\cite{pang2020deep}.
Our work does not propose new AppMult designs.
Instead, we develop a general retraining framework to recover accuracy when AccMults are replaced with existing AppMults.

\subsection{AppMult-aware Retraining Frameworks}\label{subsect:related-works-retraining}
Replacing AccMults with AppMults often reduces model accuracy.
To address this,
many frameworks retrain models with AppMults in the loop.
All prior retraining frameworks apply the \tit{straight-through estimator (STE)}.
which approximates the AppMult gradient using the AccMult gradient.
This often causes sub-optimal retraining and lower accuracy.

He~\etal{}~\cite{he2018axtrain} proposed \tit{AxTrain}, 
a Pytorch~\cite{paszke2019pytorch}-based AppMult-aware NN retraining framework.
AxTrain proposed to apply STE for gradient estimation of AppMults.
De la Parra~\etal{}~\cite{de2020proxsim} introduced \tit{ProxSim},
a TensorFlow~\cite{pang2020deep}-based NN retraining framework.
It supports integer quantization and behavior-level AppMult simulation.
Qian~\etal{}~\cite{qian2021approximate} built a C++ CPU framework that
couples AppMults synthesis with NN retraining.
It analyzes data distributions in NNs to guide AppMult synthesis,
supporting integer AppMults with LUT-based AppMult simulation.
Danopoulos~\etal{}~\cite{danopoulos2022adapt} proposed \tit{AdaPT}, a PyTorch-based NN retraining framework using integer quantization and multi-core CPU behavior-level simulation of AppMults.
Gong~\etal{}~\cite{gong2023approxtrain} presented \tit{ApproxTrain}, 
a TensorFlow-based NN retraining framework for floating-point AppMults,
also using LUT-based AppMult simulation.
Novkin~\etal{}~\cite{novkin2023approximation} propoded a PyTorch-based framework for graph NNs with integer AppMults.
Yu~\etal{}~\cite{yu2024toward} proposed a PyTorch-based retraining framework for large-scale.
They introduced a sparse integer quantization scheme
that steers the weights away from error-prone regions,
enabling AppMult behavior to be simulated with AccMults.
They also designed AppMults compatible with this scheme.
Danopoulos~\etal{}~\cite{danopoulos2025transaxx} proposed \tit{TransAxx}, 
a PyTorch-based transformer retraining framework.
It supports various bit-widths, integer quantization, LUT-based simulation for low-bit AppMults, and behavior-level simulation for high-bit AppMults.

\subsection{Layer-Wise Assignment of Heterogeneous AppMults}

Beyond single-type AppMult retraining,
several works assign different AppMults across NN layers 
to balance the model accuracy and hardware cost.
Mrazek~\etal{}~\cite{mrazek2019alwann} proposed \tit{ALWANN},
an automatic layer-wise NN approximation framework,
which uses the NSGA-II algorithm to select a layer-wise mix of different AppMults without retraining.
Guella~\etal{}~\cite{guella2024marlin} presented \tit{MARLIN},
a deployment framework for reconfigurable AppMults,
also applying NSGA-II for per-layer AppMult selection.
These methods are complementary to ours. 
Once a layer-wise assignment is chosen, 
our AppMult-aware retraining framework can be applied to recover accuracy.

\section{Preliminaries}\label{sect:prel}

This section introduces the preliminaries of AppMult-aware retraining.
We first present AppMults in subsection~\ref{subsect:appmult}
and then describe AppMult-aware retraining in subsection~\ref{subsect:gd}.

\subsection{Approximate Multipliers (AppMults)}\label{subsect:appmult}

This paper focuses on \tit{integer AppMults},
which are widely used in deep learning accelerators~\cite{simon2021exact,jain2022learning,hu2024configurable}.
In what follows, we refer to them \cadd{simply} as AppMults.
Our proposed framework supports both signed and unsigned AppMults.
Without loss of generality,
we use unsigned AppMults for illustration in the remainder of this paper.

\begin{figure}[!htbp]
    \centering
    \includegraphics[width=1.0\columnwidth]{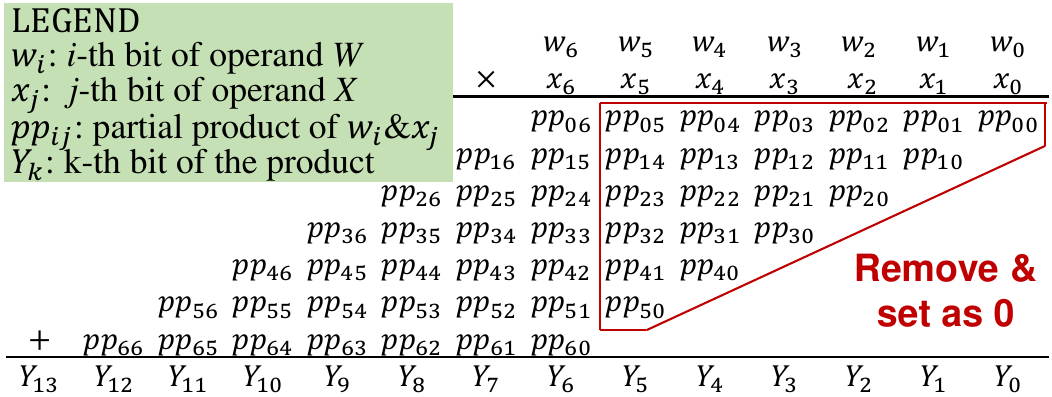}
    \caption{A simple 7-bit unsigned AppMult,
    where the rightmost 6 columns of partial products are removed.
    The figure is adapted from~\cite{hanif2019cann}.}
    \label{fig:appmult}
\end{figure}

A general AppMult with integer input operands $W$ and $X$ and integer output $Y$ implements the function:
\begin{equation}\label{eq:appmult}
    Y = \tit{AM}(W,X) = WX + \epsilon(W, X),
\end{equation}
where $WX$ is the exact product,
\tit{AM} is the AppMult function,
and $\epsilon(W, X)$ is the approximation error.
For example, Fig.~\ref{fig:appmult} shows a 7-bit unsigned AppMult,
which removes the rightmost 6 columns of partial products.
Its approximation error is expressed as
$\epsilon(W, X) = -\sum_{i=0}^{5} \sum_{j=0}^{5-i} \left(2^{i+j} pp_{ij}\right)$,
where $pp_{ij}$ is the partial product of $w_i$ (the $i$-th bit of $W$) and $x_j$ (the $j$-th bit of $X$).

To evaluate the accuracy of a $B$-bit AppMult,
common error metrics include \tit{error rate (ER)}, \tit{normalized mean error distance (NMED)}, and \tit{maximum error distance (MaxED)}~\cite{jiang2020approximate}:
\begin{equation}\label{eq:err-metrics}
\begin{aligned}
\tit{ER} & = \sum_{1\leq i \leq 2^{2B}:Y^{(i)} \neq Y_{acc}^{(i)}}{p_i}, \\
\tit{NMED} & = \sum_{i=1}^{2^{2B}}{\frac{|Y^{(i)} - Y_{acc}^{(i)}|\cdot p_i}{2^{2B} - 1}}, \\
\tit{MaxED} &= \max_{1\leq i \leq 2^{2B}}{|Y^{(i)} - Y_{acc}^{(i)}|}, \\
\end{aligned}
\end{equation}
where $Y^{(i)}$ and $Y_{acc}^{(i)}$ are the outputs of the AppMult and the \tit{accurate multiplier (AccMult)}, respectively, under the $i$-th input combination,
$p_i$ is the probability of the $i$-th input combination,
and $2^{2B}$ is the total number of input combinations.

\subsection{AppMult-Aware Retraining}\label{subsect:gd}

As shown in Fig.~\ref{fig:motivation},
AppMult-aware retraining is used to recover the accuracy of deep learning models after replacing AccMults with AppMults.
Mainstream retraining techniques rely on gradient descent and consist of two key steps: 
forward propagation and backward propagation.

During forward propagation, 
the input data passes through the model to compute the output.
In this step,
AppMults are simulated to perform approximate multiplications,
typically through \tit{lookup table (LUT)}-based methods (\eg{}, \cite{gong2023approxtrain}) or behavioral-level simulations (\eg{}, \cite{de2020proxsim}).

During the backward propagation, gradients of the loss function \wrt{} model parameters are computed,
and the parameters are updated by gradient descent. 
This step involves computing the gradients of AppMults,
and to the best of our knowledge, 
all existing AppMult-aware retraining frameworks employ the \tit{straight-through estimator (STE)}.
Specifically, these frameworks approximate the gradient of an AppMult using that of an AccMult.

For a general AppMult in Eq.~\eqref{eq:appmult},
the STE estimates its gradient as:
\begin{equation}\label{eq:ste}
\begin{aligned}
    \frac{\partial \tit{AM}}{\partial W} \approx X, \quad
    \frac{\partial \tit{AM}}{\partial X} \approx W.
\end{aligned}
\end{equation}
In other words, STE assumes that the gradients of the approximation error, $\frac{\partial\epsilon}{\partial W}$ and $\frac{\partial\epsilon}{\partial X}$, are zero.
The assumption holds when the AppMult error $\epsilon$ is small,
so its gradient closely matches that of the AccMult.
However, for an AppMult with relatively large error,
its gradient can deviate largely from the AccMult gradient.
In this case,
the STE method may yield suboptimal retraining accuracy.

To overcome this limitation,
our work proposes more precise gradient estimation methods for AppMults,
aiming to improve retraining accuracy,
particularly for relatively large-error AppMults.

\section{AppMult-Aware Retraining Framework}\label{sect:framework}

\begin{figure*}[!htbp]
    \centering
    \includegraphics[width=1.5\columnwidth]{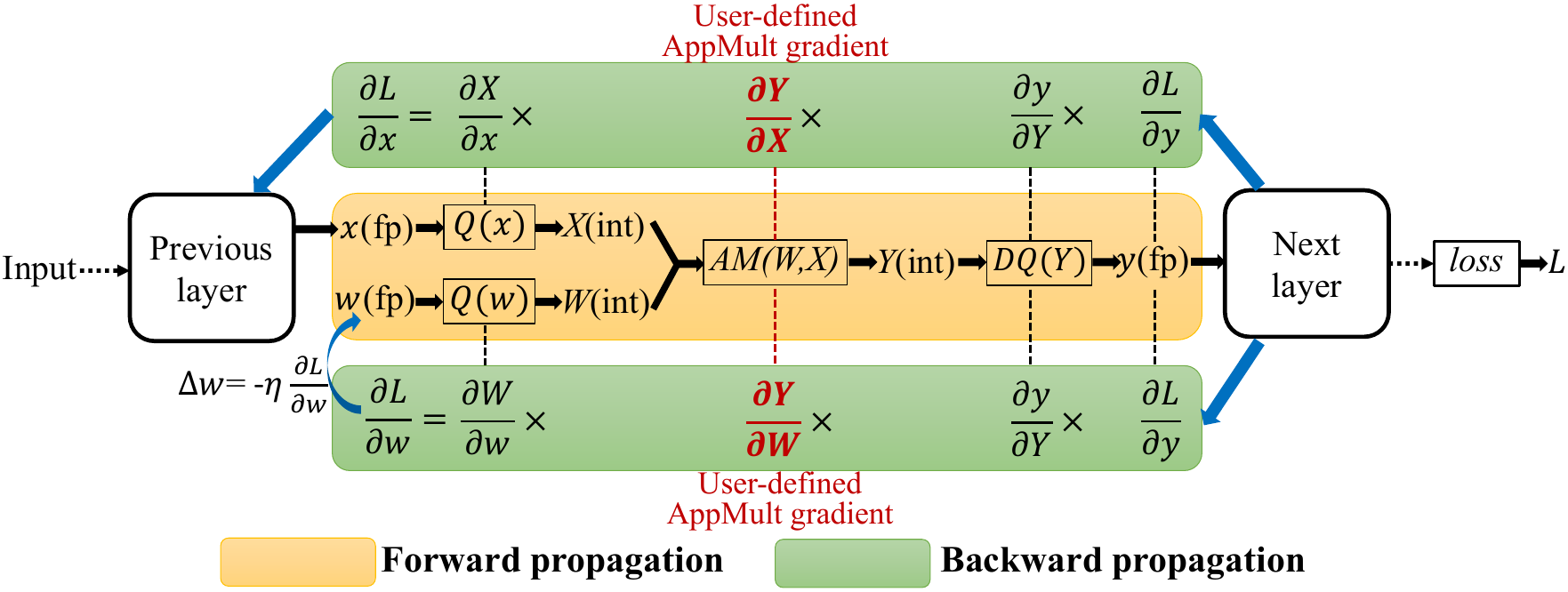}
    \caption{Forward and backward propagation in the AppMult-aware retraining framework. 
    The term ``fp'' denotes ``floating point''.
    $Q$ and $DQ$ are the quantization and dequantization functions,
    defined in Eqs.~\eqref{eq:quant} and~\eqref{eq:dequant}, respectively.
    The {\color[HTML]{c00000} red} expressions $\frac{\partial Y}{\partial X}$ and $\frac{\partial Y}{\partial W}$ represents the gradients of the AppMult function $\tit{AM}(W, X)$ \wrt{} $X$ and $W$,
    respectively.}
    \label{fig:framework}
\end{figure*}

We develop an AppMult-aware retraining framework that supports various deep learning models,
including NNs and transformer-based models.
Unlike existing frameworks described in Section~\ref{subsect:related-works-retraining},
the key feature of our framework is supporting \tit{user-defined gradients of AppMults}.
With more precise gradients than the conventional STE-based approach,
our framework achieves improved retraining accuracy.

As shown in Fig.~\ref{fig:framework},
our framework consists of forward and backward propagation.
We describe them in the following subsections.

\subsection{Forward Propagation}
During the forward propagation,
the model operates with integer AppMults.
We simulate quantization and AppMult behaviors using the method in~\cite{vaverka2020tfapprox}.

For quantization simulation,
we adopt the standard \tit{fake quantization} technique~\cite{jacob2018quantization}.
As illustrated in the middle yellow part of Fig.~\ref{fig:framework},
floating-point weight $w$ and activation $x$ are quantized into integers $W$ and $X$ through the quantization function $Q$.
For example,
a simple $B$-bit uniform asymmetric quantizer 
can be expressed  as
\begin{equation}\label{eq:quant}
\begin{aligned}
    W & = Q(w) = \text{clamp}\left(\left\lfloor\frac{w}{s_w}\right\rceil + Z_w, 0, 2^B-1\right),  \\
    X & = Q(x) = \text{clamp}\left(\left\lfloor\frac{x}{s_x}\right\rceil + Z_x, 0, 2^B-1\right)  \\
\end{aligned}
\end{equation}%
where $\lfloor \cdot \rceil$ denotes rounding,
$\tit{clamp}(a, l, u)$ restricts $a$ to the range of $[l, u]$,
$s_w$ and $s_x$ are floating-point quantization scales for weights and activations,
and $Z_w$ and $Z_x$ are integer zero points.

For asymmetric layer-wise quantization,
the quantization parameters are computed as
\begin{equation}\label{eq:observer}
\begin{aligned}
    s_w & = \frac{w_\tit{max} - w_\tit{min}}{2^B - 1}, \quad Z_w = -\left\lfloor\frac{w_\tit{min}}{s_w}\right\rceil, \\
    s_x & = \frac{x_\tit{max} - x_\tit{min}}{2^B - 1}, \quad Z_x = -\left\lfloor\frac{x_\tit{min}}{s_x}\right\rceil, \\
\end{aligned}
\end{equation}%
where $w_\tit{max}$ and $w_\tit{min}$ are the maximum and minimum weights at the current layer,
and $x_\tit{max}$ and $x_\tit{min}$ are the maximum and minimum activations at the current layer.
Other advanced quantization methods, 
such as LSQ~\cite{esser2019learned} or learnable weight clipping~\cite{shao2023omniquant},
can also be used in our framework.

After computing the AppMult function $Y=\tit{AM}(W, X)$,
the dequantization function $DQ$ converts the integer $Y$ back into floating-point $y$:
\begin{equation}\label{eq:dequant}
    y=DQ(Y) = s_w s_x(Y - Z_x W - Z_w X + Z_w Z_x).
\end{equation}%

To simulate AppMult behavior, 
\ie{}, the function $\tit{AM}(W, X)$ in the center of Fig.~\ref{fig:framework},
we utilize a LUT-based method. 
Specifically, we precompute $\tit{AM}(W, X)$ for all $(W, X)$ combinations
and store the results in LUTs.
An CUDA~\cite{cuda} kernel then implements approximate \tit{general matrix multiplication (GEMM)} by referencing the precomputed LUTs.
Since modern accelerators typically use bit-widths no more than 8~\cite{liang2021pruning},
LUTs remain compact and can be stored in GPU memory.
For example, a 7-bit AppMult LUT requires $2^{14}$ entries.
Storing these entries as 16-bit integers consumes only 32KB,
which fits into fast GPU shared memory.

\subsection{Backward Propagation}
During the backward propagation,
the gradients of the loss function $L$ \wrt{} the weight $w$ and activation $x$ are computed.
Based on the chain rule,
$\frac{\partial L}{\partial w}$ (see the bottom part of Fig.~\ref{fig:framework})
and $\frac{\partial L}{\partial x}$ (see the top part of Fig.~\ref{fig:framework})
can be computed as follows:
\begin{equation}\label{eq:chain}
\begin{aligned}
    & \frac{\partial L}{\partial w} &=& \frac{\partial W}{\partial w} {\color[HTML]{c00000} \frac{\partial Y}{\partial W}} \frac{\partial y}{\partial Y} \frac{\partial L}{\partial y} &=& Q'(w) {\color[HTML]{c00000}\frac{\partial AM}{\partial W}} DQ'(Y) \frac{\partial L}{\partial y} &, \\
    & \frac{\partial L}{\partial x} &=& \frac{\partial X}{\partial x} {\color[HTML]{c00000} \frac{\partial Y}{\partial X}} \frac{\partial y}{\partial Y} \frac{\partial L}{\partial y} &=& Q'(x) {\color[HTML]{c00000}\frac{\partial AM}{\partial X}} DQ'(Y) \frac{\partial L}{\partial y} &
\end{aligned}
\end{equation}%
Here, the computation of $Q'(w)$, $Q'(x)$, $DQ'(Y)$, and $\frac{\partial L}{\partial y}$ follows the same approach in~\cite{de2020proxsim,danopoulos2022adapt,gong2023approxtrain}.
The key difference lies in
$\frac{\partial AM}{\partial W}$ and $\frac{\partial AM}{\partial X}$,
which are defined by user-specified AppMult gradients.
Our framework accommodates arbitrary user-defined AppMult gradients represented in LUTs,
allowing convenient exploration of different gradient estimation methods during retraining.
In the next section, we propose two LUT-based methods for more precise AppMult gradient estimation,
which achieves higher retraining accuracy than the conventional STE-based approach.

\section{Gradient Estimation of AppMults for High-Accuracy Retraining}\label{sect:grad}

This section describes how to compute the gradients of AppMults,
\ie{}, $\frac{\partial \tit{AM}}{\partial W}$ and $\frac{\partial \tit{AM}}{\partial X}$ in Eq.~\eqref{eq:chain},
where $\tit{AM}$ is the AppMult function.
For clarity,
we only introduce the computation of $\frac{\partial \tit{AM}}{\partial X}$,
noting that the computation of $\frac{\partial \tit{AM}}{\partial W}$ is similar.
In what follows, we first provide an overview of existing and proposed AppMult gradient estimators in subsection~\ref{sect:grad-overview}.
We then propose two LUT-based methods in subsections~\ref{sect:difference} and~\ref{sect:grad1d}.

\subsection{Overview of Different AppMult Gradient Estimation Methods}\label{sect:grad-overview}
We propose two gradient estimation techniques,
both utilizing LUTs to store the AppMult gradients.
Based on the storage format,
we refer to them as the LUT-2D and LUT-1D methods.
In the LUT-2D method,
the AppMult gradient $\frac{\partial AM}{\partial X}$ is treated as a two-input function of integer operands $W$ and $X$.
The function $\frac{\partial AM}{\partial X}(W, X)$ is precomputed and stored in a 2D LUT,
with the first dimension indexing $W$ and the second indexing $X$.
To reduce memory footprint and improve efficiency,
the LUT-1D method represents $\frac{\partial AM}{\partial X}$ as a single-input function of $X$ only.
The function $\frac{\partial AM}{\partial X}(X)$ is stored in a 1D LUT indexed by $X$.

Table~\ref{tab:grad-comparison} compares the two proposed methods and the conventional STE-based method. 
In summary, the LUT-2D method requires the largest memory
but achieves the highest retraining accuracy, 
at the cost of the longest runtime.
The LUT-1D method offers a balanced trade-off,
achieving high retraining accuracy with moderate memory and runtime.
The STE-based method requires no LUTs
and has the smallest memory and shortest runtime,
but also the lowest retraining accuracy.
We detail the LUT-2D and LUT-1D methods in the following subsections.

\begin{table}[!htbp]
\begin{threeparttable}
\centering
\caption{Comparison of different AppMult gradient computation methods.
$B$ denotes the AppMult bit-width.}
\tabcolsep=5pt
\label{tab:grad-comparison}
\begin{tabular}{ccccc}
\toprule
\makecell{Method} & \makecell{Memory of\\AppMult Gradients} & \makecell{Retraining\\Accuracy\tnote{+}} & \makecell{Retraining\tnote{*}\\Runtime/epoch} &  \\ \midrule
LUT-2D & High ($2^{2B}$ entries)  & Highest & Long (1.9x)   &  \\
LUT-1D & Medium ($2^B$ entries) & High & Medium (1.2x) &  \\
STE    & Low (No LUT)                    & Low  & Short (1.0x)    &  \\ \bottomrule
\end{tabular}
\begin{tablenotes}
    \item[+] \cadd{Detailed retraining accuracy results are reported in Section~\ref{sect:result}}.
    \item[*] Retraining runtime was profiled on CIFAR-10~\cite{krizhevsky2009learning} with the VGG19~\cite{simonyan2014very} model using a single NVIDIA GeForce RTX 3090 GPU and batch size of 64.
    The applied AppMult is mult8u\_{1DMU} from the EvoApproxLib~\cite{mrazek2017evoapprox8b} library.
\end{tablenotes}
\end{threeparttable}
\end{table}
\subsection{LUT-2D Method: High-Accuracy AppMult Gradients}\label{sect:difference}

The LUT-2D method uses a 2D LUT,
denoted as $\tit{Grad2D}_{X}[W, X]$,
to store the gradient $\frac{\partial \tit{AM}}{\partial X}$ at each integer input pair $(W, X)$.
This method characterizes the gradient in a fine-grained way,
capturing variations in the changing rate of the AppMult function $\tit{AM}(W, X)$ near different inputs $(W, X)$.
For a $B$-bit AppMult,
$\tit{Grad2D}_{X}[W, X]$ contains $2^{2B}$ entries,
each corresponding to a unique pair $(W, X)$.

For clarity,
we illustrate the LUT-2D method using unsigned AppMults,
though it extends naturally to signed cases.
The 2D gradient LUT is computed in two steps:
1) smoothing the AppMult function, and
2) computing a difference-based gradient.
Sections~\ref{subsect:average} and \ref{subsect:gradient} describe these steps,
and Section~\ref{subsect:lut2d-cuda} explains the CUDA kernel implementation.

\subsubsection{Smoothing the AppMult Function}\label{subsect:average}

\begin{figure*}[!htbp]
\centering
\subfigure[The AppMult function,
its smoothed AppMult function using a half window size of 4,
and the AccMult function when $W_f=10$.]{\label{fig:stair}%
\includegraphics[width=0.70\columnwidth]{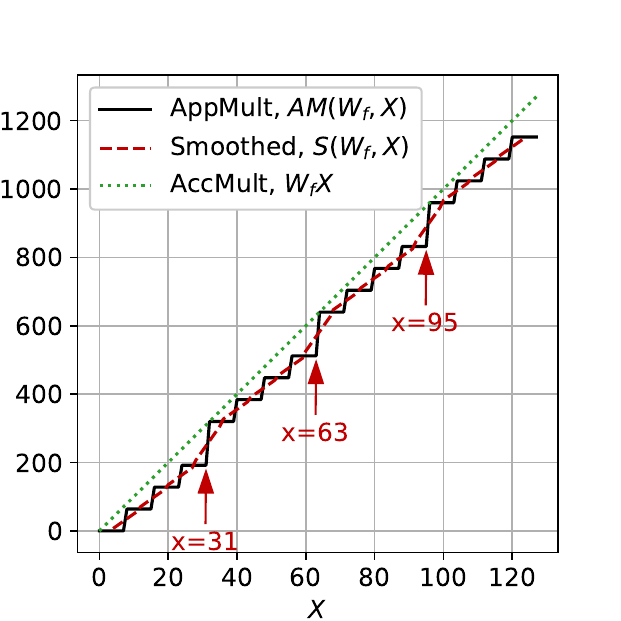}%
}
\hspace{0.2\columnwidth}
\subfigure[LUT-2D, LUT-1D, and STE-based gradients of the AppMult when $W_f=10$.]{\label{fig:grad}%
\includegraphics[width=0.70\columnwidth]{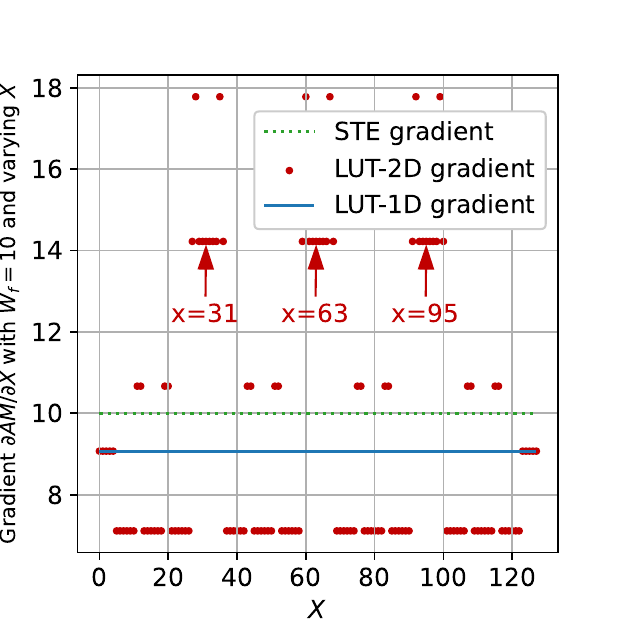}%
}
\caption{Smoothing of a 7-bit unsigned AppMult function $AM(W_f=10, X)$ and its gradient estimations.
The AppMult corresponds to Fig.~\ref{fig:appmult},
which removes the 6 rightmost columns of partial products.
The red arrows show three large changes in the AppMult function,
corresponding to the large values in the LUT-2D gradient.}
\end{figure*}

To compute $\frac{\partial \tit{AM}}{\partial X}$,
we analyze $\tit{AM}(W_f, X)$,
where $W_f$ is a fixed integer operand.
Since AppMults often approximate or remove least significant partial products,
$\tit{AM}(W_f, X)$ typically exhibit a stair-like pattern \wrt{} $X$ 
(see the black curve in Fig.~\ref{fig:stair}).
As a result,
$\frac{\partial \tit{AM}}{\partial X}$ is zero for most $X$,
and \cadd{exhibits} large values at the stair edges.
This behavior is problematic for gradient descent,
because zero gradients prohibit parameter updates,
while spikes at edges can destabilize the gradient descent process.
Thus, directly using the true gradient leads to poor retraining results.

We address this by smoothing $\tit{AM}(W_f,X)$ into $S(W_f, X)$ \cadd{via local averaging}:
\begin{equation}\label{eq:average}
\begin{aligned}
    & S(W_f, X) = \frac{1}{2\tit{HWS}+1} \sum_{\Delta x=-\tit{HWS}}^{\tit{HWS}} \tit{AM}(W_f, X+\Delta x), \\ 
    & \text{for }\tit{HWS}\le X \le 2^B-1-\tit{HWS}, \\
\end{aligned}
\end{equation}
where $B$ is the bitwidth and \tit{HWS} is a positive integer called \tit{half window size}.

For each input pair $(W_f, X)$,
Eq.~\eqref{eq:average} considers its neighbor points within a window of size $(2\tit{HWS}+1)$,
and calculates the average of the AppMult outputs in this window to produce the smoothed output $S(W_f, X)$.
For example, the red curve in Fig.~\ref{fig:stair} shows the smoothed function $S(W_f=10, X)$ 
for the AppMult function $\tit{AM}(W_f=10, X)$ (shown in blue) with $\tit{HWS}=4$.
After smoothing, $S(W_f=10, X)$ has no zero gradients and no large gradients,
making it more suitable for gradient descent.
Note that Eq.~\eqref{eq:average} is only applied to $\tit{HWS}\le X \le 2^B-1-\tit{HWS}$.
This is because only $S(W_f, X)$ for $\tit{HWS}\le X \le 2^B-1-\tit{HWS}$ is used in the LUT-2D gradient estimation,
which will be shown in Section~\ref{subsect:gradient}.

\subsubsection{Difference-Based Gradient Computation}\label{subsect:gradient}

After smoothing,
the 2D gradient LUT is computed using the following difference-based method:
\begin{equation}\label{eq:diff-x}
\begin{aligned}
    & \tit{Grad2D}_{X}[W_f, X]
    = \frac{S(W_f, X)}{\partial X} \\
    \approx & \frac{S(W_f, X+1) - S(W_f, X-1)}{2}, \\
    & \text{for }\tit{HWS} < X < 2^B-1-\tit{HWS}. \\
\end{aligned}
\end{equation}
Eq.~\eqref{eq:diff-x} considers the two neighboring points of $(W_f, X)$,
\ie{}, $(W_f, X+1)$ and $(W_f, X-1)$,
and uses the slope between them to approximate the gradient.

Note that Eq.~\eqref{eq:diff-x} is only applied to $\tit{HWS} < X < 2^B-1-\tit{HWS}$.
For the other values of $X$,
we approximate the gradient as
\begin{equation}\label{eq:diff-x-boundary}
\begin{aligned}
    & \tit{Grad2D}_{X}[W_f, X]\\ 
    = & \frac{\max\limits_{0 \le X' < 2^B} AM(W, X') - \min\limits_{0 \le X' < 2^B} AM(W, X')}{2^B - 1}, \\
    & \text{for }0 \le X \le \tit{HWS}\text{ and } 2^B-1-\tit{HWS} \le X < 2^B. \\
\end{aligned}
\end{equation}
Eq.~\eqref{eq:diff-x-boundary} computes the maximum and minimum values of $AM(W_f, X')$ for $X'\in[0, 2^B-1]$,
which is the total change of $AM(W_f, X')$ for all possible values of $X'$ under the fixed $W_f$.
This total change is then divided by
$(2^B - 1)$ to obtain the average change per unit $X$ as the gradient approximation.

\begin{example}\label{ex:lut2d}
Using Eqs.~\eqref{eq:diff-x} and~\eqref{eq:diff-x-boundary}
with $W_f=10$ and $\tit{HWS}=4$,
the LUT-2D gradient is shown in Fig.~\ref{fig:grad} in red.
For comparison,
the STE-based gradient is plotted in green in Fig.~\ref{fig:grad},
showing a constant value of 10 for all values of $X$ as $W_f=10$.
The LUT-2D gradient reflects the varing changing rate in $\tit{AM}(W_f, X)$ \wrt{} $X$,
where larger gradient values indicate greater changes of $\tit{AM}(W_f, X)$ per unit $X$.
Note that $\tit{AM}(W_f=10, X)$ has three relatively large changes at $X=31$, $63$, and $95$, 
indicated by the red arrows in Fig.~\ref{fig:stair}.
Correspondingly, the LUT-2D gradient in Fig.~\ref{fig:grad} exhibits large values near these points.
In contrast, the STE method misses these variations,
providing less informative guidance for gradient descent.
Thus, the LUT-2D method can lead to higher retraining accuracy.
\end{example}

\subsubsection{Backward Propagation with 2D Gradient LUTs}\label{subsect:lut2d-cuda}

In deep learning models,
whether NNs or a transformers,
the main operation that uses AppMults is GEMM.
During backward propagation, 
we therefore need the backward pass of approximate GEMM.

Let $\bf Y = \tit{AppGEMM(\bf W, \bf X)}$,
where the integer weight matrix $\bf W$ has size $M \times K$,
the integer activation matrix $\bf X$ has size $K \times N$.
Then, the integer output matrix $\bf Y$ has size $M \times N$,
with entries
\begin{equation}\label{eq:app-gemm}
    Y_{ij} = \sum_{k=0}^{K-1} \tit{AM}(W_{ik}, X_{kj}), 0 \le i < M, 0 \le j < N.
\end{equation}

Given the gradient of the loss function $L$ \wrt{} $Y_{ij}$,
denoted $\frac{\partial L}{\partial Y_{ij}}$,
the gradient of $L$ \wrt{} $X_{kj}$ is
\begin{equation}\label{eq:app-gemm-grad2d}
\begin{aligned}
    \frac{\partial L}{\partial X_{kj}} = & \sum_{i=0}^{M-1} \frac{\partial L}{\partial Y_{ij}} \times \frac{\partial Y_{ij}}{\partial X_{kj}}, \\ 
    = & \sum_{i=0}^{M-1} \frac{\partial L}{\partial Y_{ij}} \times \frac{\partial \tit{AM}(W_{ik}, X_{kj})}{\partial X_{kj}}, \\
    & \text{for }0 \le k < K, 0 \le j < N. \\
\end{aligned}
\end{equation}
We store the AppMult gradient in a 2D LUT
and compute $\frac{\partial \tit{AM}(W_{ik}, X_{kj})}{\partial X_{kj}}$ by reading the LUT entry $\tit{Grad2D}_{X}[W_{ik}, X_{kj}]$.

We implement a CUDA kernel for this backward pass.
Specifically,
we precompute the AppMult gradients \wrt{} $X$ for all integer pairs $(W, X)$ using Eqs.~\eqref{eq:diff-x} and~\eqref{eq:diff-x-boundary} and store them in the 2D LUT $\tit{Grad2D}_{X}$.
The precomputation of the AppMult gradient \wrt{} $W$ can be done similarly,
resulting in another 2D LUT $\tit{Grad2D}_{W}$.
The CUDA kernel receives these 2D gradient LUTs and calculates Eq.~\eqref{eq:app-gemm-grad2d}.
Since AppMults of up to 8 bits are considered,
the LUT sizes are small enough to fit in GPU memory,
making the approach efficient for retraining.

\subsection{LUT-1D Method: Memory- and Runtime-Efficient AppMult Gradients}\label{sect:grad1d}

A drawback of the LUT-2D method is its large table size of $2^{2B}$ entries.
Such large LUTs slow down gradient lookups.
To reduce memory and runtime,
we construct a more compact 1D LUT.
We first describe its construction and
then explain the backward propagation using the 1D LUT.

\subsubsection{1D Gradient LUT Construction}
We use a 1D LUT, $\tit{Grad1D}_{X}[W]$, 
to store $\frac{\partial \tit{AM}}{\partial X}$ for each integer $W$.
Here, $\tit{Grad1D}_{X}[W]$ depends only on $W$ and is independent of $X$.
Thus, $\tit{Grad1D}_{X}$ contains $2^{B}$ entries,
one for each possible integer $W$.
Each entry is computed as
\begin{equation}\label{eq:grad1d}
\begin{aligned}
    &\tit{Grad1D}_{X}[W] \\
    = & \frac{\max\limits_{0 \le X' < 2^B} AM(W, X') - \min\limits_{0 \le X' < 2^B} AM(W, X')}{2^B - 1}. \\
\end{aligned}
\end{equation}
This corresponds to the average changing rate of $AM(W, X)$ acroos all possible $X$ values at a fixed $W$.

The LUT-1D method can be viewed as a special case of the LUT-2D method.
When the HWS value in the LUT-2D method is $2^{B-1} - 1$,
Eq.~\eqref{eq:diff-x} is never used,
and Eq.~\eqref{eq:diff-x-boundary} applies for all $X$ at a fixed $W$.
In this case,
the LUT-2D method reduces exactly to the LUT-1D method.

\begin{example}
Consider the same AppMult function $\tit{AM}(W_f=10,X)$ as in Example~\ref{ex:lut2d}.
The LUT-1D gradient from Eq.~\eqref{eq:grad1d} is shown in Fig.~\ref{fig:grad} with the blue line.
Comparing with the STE-based gradient in green,
which is a constant 10,
the LUT-1D method yieads a smaller gradient value of 9.07.
This is reasonable since the AppMult function in Fig.~\ref{fig:stair} grows more slowly than the AccMult function,
whose gradient equals 10 when $W_f=10$.
Although the LUT-1D method does not capture fine-grained variations as the LUT-2D method does,
it reduces memory and lookup time.
It also provides a more precise gradient estimation than STE,
potentially improving retraining accuracy.
\end{example}

\subsubsection{Backward Propagation with 1D Gradient LUTs}
The backward propagation for AppGEMM with the 1D gradient LUTs follows the procedure in Section~\ref{subsect:lut2d-cuda}.
The only difference is that in Eq.~\eqref{eq:app-gemm-grad2d},
the gradient term 
$\frac{\partial \tit{AM}(W_{ik}, X_{kj})}{\partial X_{kj}}$
is obtained from the 1D LUT entry $\tit{Grad1D}_{X}[W_{ik}]$,
instead of the 2D LUT entry.
Since the 1D LUT has only has $2^{B}$ entries,
gradient lookups are much faster than with the $2^{2B}$-entry 2D LUT,
leading to shorter retraining runtime.

\section{Experimental Results}\label{sect:result}

We implemented our AppMult-aware retraining framework in PyTorch 2.4 and CUDA 12.4.
The framework supports AppMults with bit-widths up to 8 and allows user-defined gradients.
Since 8-bit and sub-8-bit arithmetic are commonly used in low-power deep learning accelerators~\cite{liang2021pruning},
we do not study AppMults beyond 8 bits.

We first report the hardware cost of the AppMults used in our experiments (Section~\ref{subsect:appmult-hardware}).
We then present the AppMult-aware retraining results on deep learning models, 
covering CNNs (Section~\ref{subsect:res-nn}) and vision transformers (Section~\ref{subsect:res-vit}).
Finally, we evaluate the effectiveness of the proposed gradient estimation methods under low-bit quantization (Section~\ref{subsect:res-extreme}).

\subsection{Hardware Cost of AppMults}\label{subsect:appmult-hardware}

\begin{table}[!htbp]
\centering
\caption{Information of tested multipliers,
including area, delay, power, and error metrics.
The term ``\_$\tit{rmk}$'' indicates the removal of the rightmost $k$ columns of partial products.
The term ``\_$\tit{syn}$'' means the AppMult is generated by the approximate logic synthesis tool~\cite{meng2020alsrac}.}\label{tab:appmult}
\tabcolsep=4pt
\begin{tabular}{lrrrrrr}
\toprule
\makecell{Unsigned\\Multipliers} & \makecell{Area\\($\mu m^2$)}  & \makecell{Delay\\($ps$)} & \makecell{Power\\($\mu W$)} & ER/\% & NMED/\% & MaxED \\ \midrule
mul8u\_acc          & 25.6  & 730.1 & 22.93 & 0.0   & 0.00    & 0     \\
mul8u\_syn1         & 13.0  & 582.2 & 9.68  & 99.1  & 0.28    & 1937  \\
mul8u\_syn2         & 12.3  & 577.7 & 9.29  & 99.5  & 0.30    & 2057  \\
mul8u\_2NDH         & 10.0  & 512.6 & 6.48  & 98.7  & 0.44    & 2709  \\
mul8u\_17C8         & 7.7   & 624.4 & 5.01  & 99.0  & 0.56    & 1577  \\
mul8u\_1DMU         & 15.6  & 837.6 & 11.09 & 66.0  & 0.65    & 4084  \\
mul8u\_17R6         & 6.9   & 743.3 & 4.60  & 99.0  & 0.67    & 1925  \\
mul8u\_rm8          & 11.6  & 655.0 & 9.19  & 98.0  & 0.68    & 1793  \\ \midrule
mul7u\_acc          & 19.0  & 695.0 & 15.72 & 0.0   & 0.00    & 0     \\
mul7u\_06Q          & 10.6  & 861.9 & 7.90  & 95.4  & 0.24    & 162   \\
mul7u\_073          & 11.0  & 889.8 & 8.61  & 95.2  & 0.27    & 154   \\
mul7u\_rm6          & 11.4  & 599.0 & 9.00  & 93.8  & 0.49    & 321   \\
mul7u\_syn1         & 11.5  & 561.3 & 9.06  & 97.6  & 0.28    & 457   \\
mul7u\_syn2         & 10.9  & 532.4 & 7.98  & 98.8  & 0.39    & 713   \\
mul7u\_081          & 10.7  & 673.6 & 7.67  & 97.3  & 0.45    & 314   \\
mul7u\_08E          & 8.9   & 612.5 & 6.15  & 97.5  & 0.46    & 317   \\ \midrule
mul6u\_acc          & 14.1  & 680.1 & 10.47 & 0.0   & 0.00    & 0     \\
mul6u\_rm4          & 10.3  & 563.9 & 7.06  & 81.3  & 0.30    & 49    \\ \midrule
mul4u\_acc          & 5.2   & 376.3 & 3.03  & 0.0   & 0.00    & 0     \\
mul4u\_rm2          & 4.3   & 309.3 & 2.40  & 50.0   & 0.49  & 5     \\ \midrule\midrule
\makecell{Signed\\Multipliers} & \makecell{Area\\($\mu m^2$)}  & \makecell{Delay\\($ps$)} & \makecell{Power\\($\mu W$)} & ER/\% & NMED/\% & MaxED \\ \midrule 
mul8s\_acc          & 26.52 & 863.3 & 25.47 & 0.0   & 0.00    & 0     \\
mul8s\_1KV9         & 22.26 & 934.2 & 23.33 & 68.8  & 0.0065  & 17    \\
mul8s\_1L2H         & 17.63 & 743.7 & 17.12 & 74.6  & 0.081   & 255   \\
mul8s\_1L2L         & 12.68 & 643.8 & 11.42 & 93.2  & 0.23    & 759   \\ \bottomrule
\end{tabular}
\end{table}


We used Synopsys Design Compiler~\cite{synopsys2024synopsys}  with the ASAP 7nm standard cell library~\cite{clark2016asap7} to measure area, delay, and power.
Power was measured at 1GHz under a uniform input distribution.
The ER, NMED, and MaxED error metrics of AppMults were measured by enumerating all input combinations under a uniform distribution,
following Eq.~\eqref{eq:err-metrics}.
We use the open-source tool in~\cite{meng2024vacsem} to obtain the ER and NMED metrics.

Table~\ref{tab:appmult} reports the tested multipliers and their area, delay, power, and error metrics (ER, NMED, and MaxED, see Eq.~\eqref{eq:err-metrics}).
They include both unsigned multipliers with bit widths of 8, 7, 6, and 4,
and signed multipliers with bit width of 8.
They are organized into three categories:
1) Simple AppMults formed by removing least-significant partial products like the one in Fig.~\ref{fig:appmult} 
(denoted ``\_rm$k$'', meaning the removal of the rightmost $k$ columns of partial products);
2) AppMults from the approximate arithmetic library EvoApproxLib~\cite{mrazek2017evoapprox8b};
and 3) AppMults generated by an approximate logic synthesis tool~\cite{meng2020alsrac} (denoted ``\_syn'').
Table~\ref{tab:appmult} also lists the information of accurate multipliers (denoted ``\_acc'') for reference.

\begin{table*}[!htbp]
\centering
\caption{Retraining results with the STE-based gradient, our LUT-1D gradient, and our LUT-2D gradient on the CIFAR-10 dataset.
The accuracies of quantized CNNs using accurate multipliers (marked with \tit{``\_acc''}) after quantization-aware training are also reported for reference.
Power consumption and delay are normalized to those of the 8-bit AccMult (\tit{mul8u\_acc}).
{\color[HTML]{00B050} \textbf{Green bold}} entries denote that our LUT-1D method outperforms the other two methods.
{\color[HTML]{0070C0} \textbf{Blue bold}} entries denote that our LUT-2D method outperforms the other two methods.%
}\label{tab:cifar10}
\begin{tabular}{c|c|rrrrrr|rrr}
\bottomrule
                            &                              &                                          & \multicolumn{5}{c|}{Accuracy after retraining}                                                                                                                                       & \multicolumn{3}{c}{Multiplier information} \\
\multirow{-2}{*}{CNN}       & \multicolumn{1}{c|}{\multirow{-2}{*}{Multiplier}} & \multicolumn{1}{c}{\multirow{-2}{*}{\makecell{Initial\\Accuracy}}} & STE     & LUT-1D                                  & \makecell{LUT-1D\\impr.\\than STE}                   & LUT-2D                                  & \makecell{LUT-2D\\impr.\\than STE}                   & \makecell{Norm.\\Power}    & \makecell{Norm.\\Delay}   & \makecell{NMED\\/\%}   \\\hline
                            & mul8u\_acc                                       & \multicolumn{6}{c|}{Reference accuracy: 92.48\%}                                                                                                                                                                                                  & 1.00           & 1.00          & 0.00      \\
                            & mul8u\_syn1                                      & 7.39 \%                                                        & 79.44\% & {\color[HTML]{00B050} \textbf{83.19\%}} & {\color[HTML]{00B050} \textbf{3.75\%}}  & 81.96\%                                 & 2.52\%                                  & 0.42           & 0.80          & 0.28      \\
                            & mul8u\_syn2                                      & 8.19 \%                                                        & 67.12\% & 82.22\%                                 & 15.10\%                                 & {\color[HTML]{0070C0} \textbf{85.94\%}} & {\color[HTML]{0070C0} \textbf{18.82\%}} & 0.41           & 0.79          & 0.30      \\
                            & mul8u\_2NDH                                      & 9.36 \%                                                        & 90.66\% & 90.82\%                                 & 0.16\%                                  & {\color[HTML]{0070C0} \textbf{91.02\%}} & {\color[HTML]{0070C0} \textbf{0.36\%}}  & 0.28           & 0.70          & 0.44      \\
                            & mul8u\_17C8                                      & 9.21 \%                                                        & 85.67\% & {\color[HTML]{00B050} \textbf{87.77\%}} & {\color[HTML]{00B050} \textbf{2.10\%}}  & 86.89\%                                 & 1.22\%                                  & 0.22           & 0.86          & 0.56      \\
                            & mul8u\_1DMU                                      & 8.84 \%                                                        & 70.29\% & 78.25\%                                 & 7.96\%                                  & {\color[HTML]{0070C0} \textbf{81.31\%}} & {\color[HTML]{0070C0} \textbf{11.02\%}} & 0.48           & 1.15          & 0.65      \\
                            & mul8u\_17R6                                      & 9.75 \%                                                        & 80.45\% & 83.98\%                                 & 3.53\%                                  & {\color[HTML]{0070C0} \textbf{85.68\%}} & {\color[HTML]{0070C0} \textbf{5.23\%}}  & 0.20           & 1.02          & 0.67      \\
                            & mul8u\_rm8                                       & 10.37\%                                                        & 63.81\% & {\color[HTML]{00B050} \textbf{75.14\%}} & {\color[HTML]{00B050} \textbf{11.33\%}} & 69.34\%                                 & 5.53\%                                  & 0.40           & 0.90          & 0.68      \\\cline{2-11}
                            & mul7u\_acc                                       & \multicolumn{6}{c|}{Reference accuracy: 92.10\%}                                                                                                                                                                                                  & 0.69           & 0.95          & 0.00      \\
                            & mul7u\_06Q                                       & 67.07\%                                                        & 91.20\% & 91.77\%                                 & 0.57\%                                  & {\color[HTML]{0070C0} \textbf{91.69\%}} & {\color[HTML]{0070C0} \textbf{0.49\%}}  & 0.34           & 1.18          & 0.24      \\
                            & mul7u\_073                                       & 83.59\%                                                        & 91.29\% & 91.44\%                                 & 0.15\%                                  & {\color[HTML]{0070C0} \textbf{91.46\%}} & {\color[HTML]{0070C0} \textbf{0.17\%}}  & 0.38           & 1.22          & 0.27      \\
                            & mul7u\_rm6                                       & 8.17 \%                                                        & 75.82\% & {\color[HTML]{00B050} \textbf{88.36\%}} & {\color[HTML]{00B050} \textbf{12.54\%}} & 87.61\%                                 & 11.79\%                                 & 0.39           & 0.82          & 0.28      \\
                            & mul7u\_syn1                                      & 8.24 \%                                                        & 83.03\% & {\color[HTML]{00B050} \textbf{88.83\%}} & {\color[HTML]{00B050} \textbf{5.80\%}}  & 88.11\%                                 & 5.08\%                                  & 0.40           & 0.77          & 0.28      \\
                            & mul7u\_syn2                                      & 10.98\%                                                        & 66.66\% & 69.16\%                                 & 2.50\%                                  & {\color[HTML]{0070C0} \textbf{71.53\%}} & {\color[HTML]{0070C0} \textbf{4.87\%}}  & 0.35           & 0.73          & 0.39      \\
                            & mul7u\_081                                       & 9.84 \%                                                        & 84.82\% & 85.78\%                                 & 0.96\%                                  & {\color[HTML]{0070C0} \textbf{87.20\%}} & {\color[HTML]{0070C0} \textbf{2.38\%}}  & 0.33           & 0.92          & 0.45      \\
                            & mul7u\_08E                                       & 71.86\%                                                        & 89.74\% & 89.68\%                                 & -0.06\%                                 & {\color[HTML]{0070C0} \textbf{90.18\%}} & {\color[HTML]{0070C0} \textbf{0.44\%}}  & 0.27           & 0.84          & 0.46      \\\cline{2-11}
\multirow{-17}{*}{VGG19}    & \makecell{VGG mean over\\7\&8-bit AppMults}      & 23.06\%                                                        & 80.00\%              & 84.74\%                                 & 4.74\%                                  & {\color[HTML]{0070C0} \textbf{84.99\%}} & {\color[HTML]{0070C0} \textbf{4.99\%}}  &                &               &           \\\hline\hline
                            & mul8u\_acc                                       & \multicolumn{6}{c|}{Reference accuracy: 93.73\%}                                                                                                                                                                                                  & 1.00           & 1.00          & 0.00      \\
                            & mul8u\_syn1                                      & 9.99 \%                                                        & 88.66\% & 91.97\%                                 & 3.31\%                                  & {\color[HTML]{0070C0} \textbf{91.42\%}} & {\color[HTML]{0070C0} \textbf{2.76\%}}  & 0.42           & 0.80          & 0.28      \\
                            & mul8u\_syn2                                      & 10.19\%                                                        & 87.67\% & 87.99\%                                 & 0.32\%                                  & {\color[HTML]{0070C0} \textbf{88.67\%}} & {\color[HTML]{0070C0} \textbf{1.00\%}}  & 0.41           & 0.79          & 0.30      \\
                            & mul8u\_2NDH                                      & 27.47\%                                                        & 92.72\% & {\color[HTML]{00B050} \textbf{93.03\%}} & {\color[HTML]{00B050} \textbf{0.31\%}}  & 92.93\%                                 & 0.21\%                                  & 0.28           & 0.70          & 0.44      \\
                            & mul8u\_17C8                                      & 12.48\%                                                        & 91.47\% & 91.50\%                                 & 0.03\%                                  & {\color[HTML]{0070C0} \textbf{92.44\%}} & {\color[HTML]{0070C0} \textbf{0.97\%}}  & 0.22           & 0.86          & 0.56      \\
                            & mul8u\_1DMU                                      & 9.94 \%                                                        & 82.95\% & 90.64\%                                 & 7.69\%                                  & {\color[HTML]{0070C0} \textbf{91.34\%}} & {\color[HTML]{0070C0} \textbf{8.39\%}}  & 0.48           & 1.15          & 0.65      \\
                            & mul8u\_17R6                                      & 12.24\%                                                        & 91.09\% & 91.13\%                                 & 0.04\%                                  & {\color[HTML]{0070C0} \textbf{91.56\%}} & {\color[HTML]{0070C0} \textbf{0.47\%}}  & 0.20           & 1.02          & 0.67      \\
                            & mul8u\_rm8                                       & 9.92 \%                                                        & 73.15\% & {\color[HTML]{00B050} \textbf{90.68\%}} & {\color[HTML]{00B050} \textbf{17.53\%}} & 90.60\%                                 & 17.45\%                                 & 0.40           & 0.90          & 0.68      \\\cline{2-11}
                            & mul7u\_acc                                       & \multicolumn{6}{c|}{Reference accuracy: 93.61\%}                                                                                                                                                                                                  & 0.69           & 0.95          & 0.00      \\
                            & mul7u\_06Q                                       & 91.30\%                                                        & 93.33\% & 93.33\%                                 & 0.00\%                                  & 93.27\%                                 & -0.06\%                                 & 0.34           & 1.18          & 0.24      \\
                            & mul7u\_073                                       & 90.77\%                                                        & 93.18\% & {\color[HTML]{00B050} \textbf{93.35\%}} & {\color[HTML]{00B050} \textbf{0.17\%}}  &                               93.34\%   &                               0.16\%    & 0.38           & 1.22          & 0.27      \\
                            & mul7u\_rm6                                       & 9.54 \%                                                        & 89.68\% & {\color[HTML]{00B050} \textbf{92.84\%}} & {\color[HTML]{00B050} \textbf{3.16\%}}  & 91.82\%                                 & 2.14\%                         & 0.39           & 0.82          & 0.28      \\
                            & mul7u\_syn1                                      & 10.15\%                                                        & 92.10\% & 92.27\%                                 & 0.17\%                                  & {\color[HTML]{0070C0} \textbf{92.89\%}} & {\color[HTML]{0070C0} \textbf{0.79\%}}  & 0.40           & 0.77          & 0.28      \\
                            & mul7u\_syn2                                      & 10.00\%                                                        & 85.78\% & {\color[HTML]{00B050} \textbf{90.96\%}} & {\color[HTML]{00B050} \textbf{5.18\%}}  & 88.32\%                                 & 2.54\%                                  & 0.35           & 0.73          & 0.39      \\
                            & mul7u\_081                                       & 12.62\%                                                        & 92.39\% & 92.42\%                                 & 0.03\%                                  & {\color[HTML]{0070C0} \textbf{92.71\%}} & {\color[HTML]{0070C0} \textbf{0.32\%}}  & 0.33           & 0.92          & 0.45      \\
                            & mul7u\_08E                                       & 86.81\%                                                        & 92.27\% & 92.17\%                                 & -0.10\%                                 & {\color[HTML]{0070C0} \textbf{92.38\%}} & {\color[HTML]{0070C0} \textbf{0.11\%}}  & 0.27           & 0.84          & 0.46      \\\cline{2-11}
\multirow{-17}{*}{ResNet18} & \makecell{ResNet mean over\\7\&8-bit AppMults}   & 28.82\%                                                      & 89.03\%              & {\color[HTML]{00B050} \textbf{91.73\%}} & {\color[HTML]{00B050} \textbf{2.70\%}}  & 91.69\%                      & 2.66\%                                  &                &               &           \\\hline\hline
\multicolumn{2}{c|}{Mean over all AppMults \& CNNs}                            & 25.94\%                                                      & 84.52\%              & 88.24\%                                 & 3.72\%                                  & {\color[HTML]{0070C0} \textbf{88.34\%}} & {\color[HTML]{0070C0} \textbf{3.83\%}}  &                \\\toprule
\end{tabular}
\end{table*}

\subsection{Experiments on AppMult-Based CNNs and CIFAR-10/100 Dataset}\label{subsect:res-nn}

This set of experiments evaluates the effectiveness of the proposed LUT-2D and LUT-1D gradient estimation methods,
compared to the STE-based method in previous works~\cite{he2018axtrain,de2020proxsim,qian2021approximate,danopoulos2022adapt,gong2023approxtrain,novkin2023approximation,yu2024toward,danopoulos2025transaxx}.
For a fair comparison, we do not directly use the existing AppMult-aware retraining frameworks.
Instead, 
we re-implemented the STE baseline in our framework.
All runs in this set of experiments use a single NVIDIA GeForce RTX 3090 GPU.

Unless otherwise specified,
the training settings are as follows:
batch size 64, number of epochs 30, Adam optimizer, 
and a learning rate schedule of 0.001 for epochs 1--10, 
0.0005 for epochs 11--20, 
and 0.00025 for epochs 21--30.
The half window size (\tit{HWS}, see Section~\ref{subsect:average}) is set to 32 for 8-bit AppMults and 16 for 7-bit AppMults,
an empirical choice that works well for the tested AppMults.
We apply layer-wise asymmetric quantization using Eq.~\eqref{eq:observer} for both weights and activations in all the linear and convolutional layers.

We evaluate our framework on the CIFAR-10 and CIFAR-100~\cite{krizhevsky2009learning} datasets
with the VGG19~\cite{simonyan2014very} and ResNet~\cite{he2016deep} CNN models.
As convolutional layers dominate the compute cost in these CNNs,
we replaced all AccMults in convolutional layers with the same type of AppMults to reduce hardware cost,
following prior practice~\cite{yu2024toward,hu2024configurable}.
In the following,
we first present the results on CIFAR-10, followed by CIFAR-100.

\subsubsection{Experiments on the CIFAR-10 Dataset}\label{subsect:res-cifar10}
This experiment compares our LUT-2D and LUT-1D methods with the STE method on the CIFAR-10 dataset using the VGG19 and ResNet18 models.
The results are shown in Table~\ref{tab:cifar10}.
The top part of Table~\ref{tab:cifar10} presents the accuracy after retraining with various 7-bit and 8-bit AppMults using the VGG19 model,
and the bottom part presents the results using the ResNet18 model.
The accuracies of the quantized CNNs using the AccMults after quantization-aware training~\cite{jacob2018quantization},
the initial accuracies using AppMults before AppMult-aware retraining,
and the power consumption, delay, and NMED of the multipliers are also reported.
Note that the power consumption and delay are normalized to those of the 8-bit AccMult (mul8u\_acc).
The green bold entries denote that our LUT-1D method outperforms the other two methods,
and the blue bold entries denote that our LUT-2D method outperforms the other two methods.

\paragraph{Comparison Using the VGG19 Model}

From the top part of Table~\ref{tab:cifar10},
for the VGG19 model,
we can see that our LUT-2D method consistently outperforms STE for all tested 8-bit and 7-bit AppMults,
and our LUT-1D method outperforms STE for most tested AppMults except for mul7u\_08E.
Compared to STE,
the LUT-1D method improves the retraining accuracy by an average of 4.74\%,
and the LUT-2D method improves the retraining accuracy by an average of 4.99\%.
This is reasonable because the LUT-1D and LUT-2D methods provide more accurate gradient estimation than STE.
Compared to the initial accuracy before AppMult-aware retraining,
our method recovers the average accuracy from 23.06\% to 84.74\% with the LUT-1D method,
and to 84.99\% with the LUT-2D method.
Notably, for mul8u\_syn2, mul8u\_rm8, and mul7u\_rm6,
LUT-1D improves the accuracy by 15.10\%, 11.33\%, and 12.54\%, respectively,
compared to STE.
For mul8u\_syn2, mul8u\_1DMU, and mul7u\_rm6,
LUT-2D improves the accuracy by 18.82\%, 11.02\%, and 11.79\%, respectively,
compared to STE.
Moreover, for mul7u\_06Q,
our method recovers the accuracy from 67.07\% to 91.89\%,
and the final accuracy is close to the reference accuracy of 92.10\% with 7-bit AccMult.
Meanwhile, mul7u\_06Q (normalized power=0.34) reduces power consumption by 51\% compared to the 7-bit AccMult (normalized power=0.69),
offering an attractive trade-off between power consumption and accuracy.

\begin{figure}[!htbp]
\centering
\subfigure[7-bit AppMults.]{\label{fig:tradeoff_vgg19_mul7}%
\includegraphics[width=0.49\columnwidth]{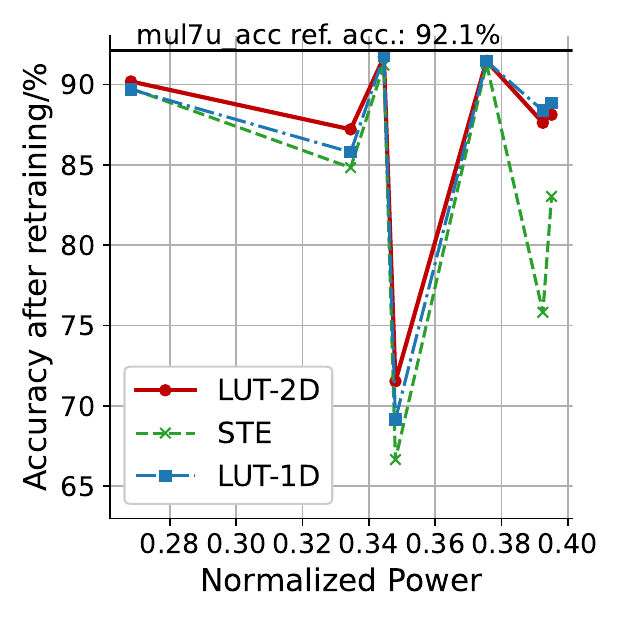}%
}\hspace{0.015\columnwidth}%
\subfigure[8-bit AppMults.]{\label{fig:tradeoff_vgg19_mul8}%
\includegraphics[width=0.49\columnwidth]{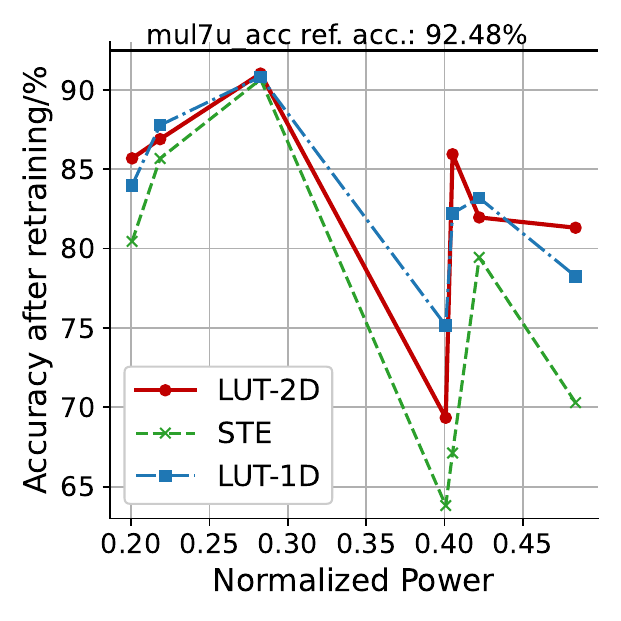}%
}
\caption{Trade-off between VGG19 accuracy and power consumption using 7-bit and 8-bit AppMults on the CIFAR-10 dataset. Power is normalized to that of the 8-bit AccMult (mul8u\_acc).}\label{fig:tradeoff_vgg19}
\end{figure}

Using the data from the top part of Table~\ref{tab:cifar10},
we plot the VGG19 accuracy after retraining versus power consumption for 7-bit and 8-bit AppMults in Fig.~\ref{fig:tradeoff_vgg19_mul7} and Fig.~\ref{fig:tradeoff_vgg19_mul8}, respectively.
In both figures, 
at the same normalized power level,
our LUT-2D method consistently outperforms the STE method in accuracy,
and our LUT-1D method outperforms STE in most cases.
Reference accuracies for the 7-bit and 8-bit AccMults are indicated by the black horizontal lines.
Our method achieves better accuracy-power trade-offs, 
with acceptable drops in accuracy compared to the reference accuracies.
In comparison, the STE method sometimes suffers from large accuracy drops.

Regarding GPU memory usage,
for the VGG19 model with the 7-bit AppMults,
the STE, LUT-1D, and LUT-2D methods all use about 2876.6 MB of GPU memory.
Note that the LUTs storing the AppMult functions and gradients are small and negligible compared to the model parameters and intermediate activations.
The LUT-1D and LUT-2D methods use slightly more memory than STE due to the storage of the lookup tables.
As for runtime,
the STE, LUT-1D and LUT-2D methods take about 0.71, 0.87, and 1.37 hours to retrain the VGG19 model with a 7-bit AppMult using a single NVIDIA RTX 3090 GPU, respectively.
The runtime overhead is primarily due to the additional computation of the LUT-based gradient during the backward propagation.
However, it is acceptable given the accuracy improvement.

\begin{figure}[!htbp]
\centering
\subfigure[7-bit AppMults.]{\label{fig:tradeoff_mul7}%
\includegraphics[width=0.49\columnwidth]{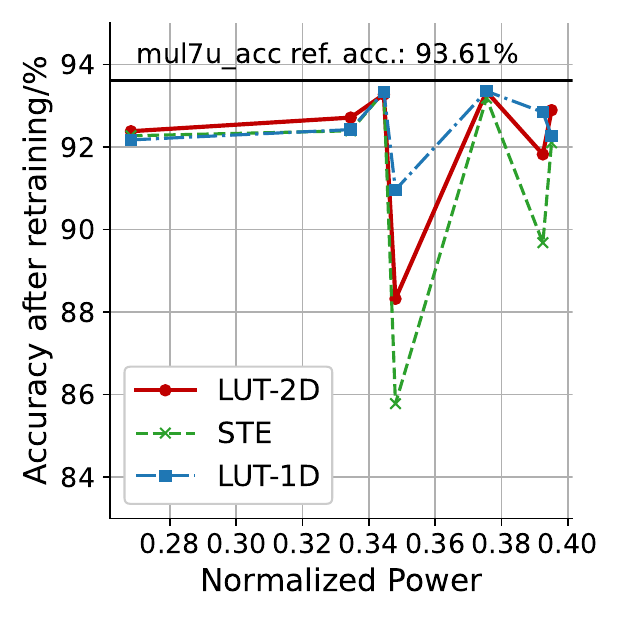}%
}\hspace{0.015\columnwidth}%
\subfigure[8-bit AppMults.]{\label{fig:tradeoff_mul8}%
\includegraphics[width=0.49\columnwidth]{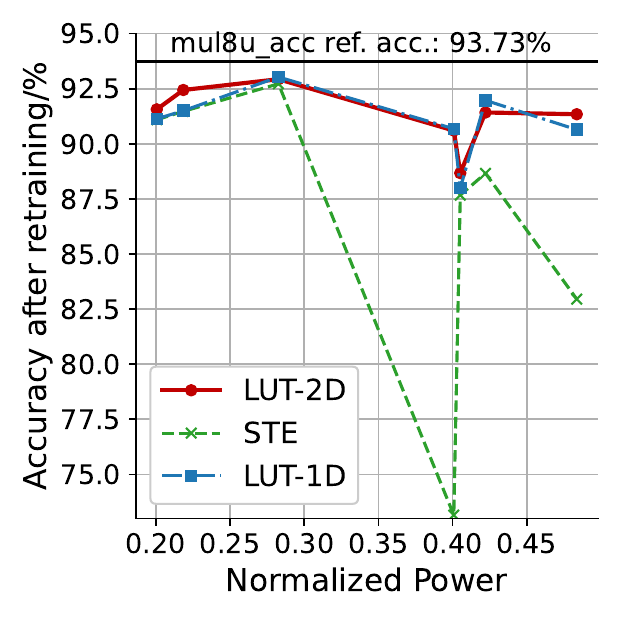}%
}
\caption{Trade-off between ResNet18 accuracy and power consumption using 7-bit and 8-bit AppMults on the CIFAR-10 dataset. Power is normalized to that of the 8-bit AccMult (mul8u\_acc).}\label{fig:tradeoff}
\end{figure}

\paragraph{Comparison Using the ResNet18 Model}
The bottom part of Table~\ref{tab:cifar10} presents the results using the ResNet18 model.
We observe that for the ResNet18 model,
our LUT-1D method achieves higher accuracy than STE for most tested 8-bit and 7-bit AppMults except for mul7u\_06Q and mul7u\_08E,
and our LUT-2D method achieves higher accuracy than STE for most AppMults except for mul7u\_06Q.
Compared to STE,
the LUT-1D method improves the retraining accuracy by an average of 2.70\%,
and the LUT-2D method improves the retraining accuracy by an average of 2.66\%.
This is again because the LUT-1D and LUT-2D methods provide more accurate gradient estimates than STE.
Compared to the accuracy before AppMult-aware retraining,
the LUT-1D method recovers the accuracy from 28.82\% to 91.73\% on average,
and the LUT-2D method recovers the accuracy from 28.82\% to 91.69\% on average.
Notably, for mul8u\_rm8,
our LUT-1D method improves the accuracy by 17.53\% compared to STE,
and our LUT-2D method improves the accuracy by 17.45\% compared to STE.
Moreover, for mul7u\_073,
our LUT-1D method achieves an accuracy of 93.35\%,
which is very close to the reference accuracy of 93.61\% for 7-bit AccMult,
while reducing the power by 45\%.
This shows that AppMults can be used in CNNs with very small accuracy loss.

Using the data from the bottom part of Table~\ref{tab:cifar10},
we plot the ResNet18 accuracy after retraining versus power consumption for 7-bit and 8-bit AppMults in Fig.~\ref{fig:tradeoff_mul7} and Fig.~\ref{fig:tradeoff_mul8}, respectively.
In both figures, 
at the same normalized power level,
our LUT-1D and LUT-2D methods outperform the STE method in accuracy in most cases.
Reference accuracies for the 7-bit and 8-bit AccMults are indicated by the black horizontal lines.
Our method achieves better accuracy-power trade-offs, 
with acceptable drops in accuracy compared to the reference accuracies.
In comparison, the STE method sometimes suffers from large accuracy drops,
sometimes reducing the accuracy by over 20\% compared to the reference accuracies for 8-bit AppMults.

Regarding GPU memory usage,
for the ResNet18 model with the 7-bit AppMults,
the STE, LUT-1D, and LUT-2D methods all use about 3208.6 MB of GPU memory.
Note that the LUTs storing the AppMult functions and gradients are small and negligible compared to the model parameters and intermediate activations.
As for runtime,
the STE, LUT-1D, and LUT-2D methods take about 0.94, 1.32, and 2.41 hours to retrain the ResNet18 model with a 7-bit AppMult using a single NVIDIA RTX 3090 GPU, respectively.
The runtime overhead is primarily due to the additional computation of the LUT-based gradient during the backward propagation.
However, it is acceptable given the accuracy improvement.

\subsubsection{Experiments on the CIFAR-100 Dataset}

\begin{figure}
\centering
\subfigure[Testing accuracy of ResNet34 versus epochs.]{\label{fig:resnet34}%
\includegraphics[width=0.48\columnwidth]{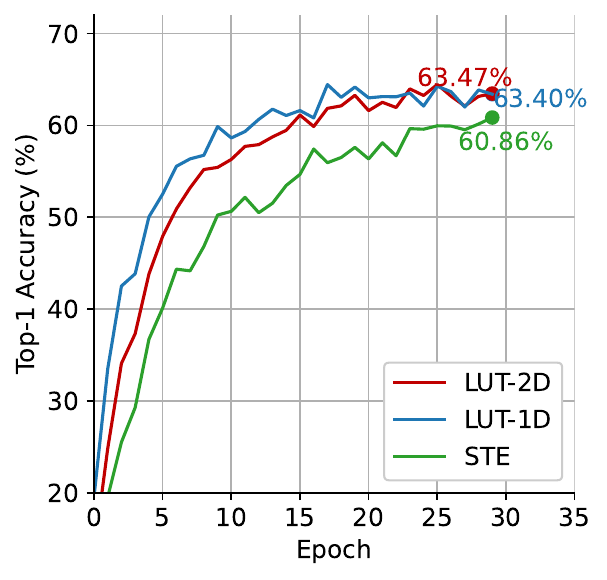}%
}\hspace{0.02\columnwidth}%
\subfigure[Testing accuracy of ResNet50 versus epochs.]{\label{fig:resnet50}%
\includegraphics[width=0.48\columnwidth]{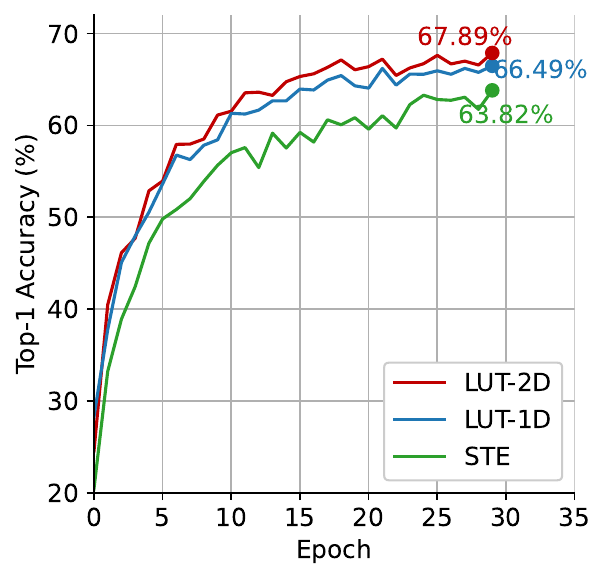}%
}

\subfigure[Testing accuracy of ResNet34 versus runtime.]{\label{fig:resnet34_time}%
\includegraphics[width=0.48\columnwidth]{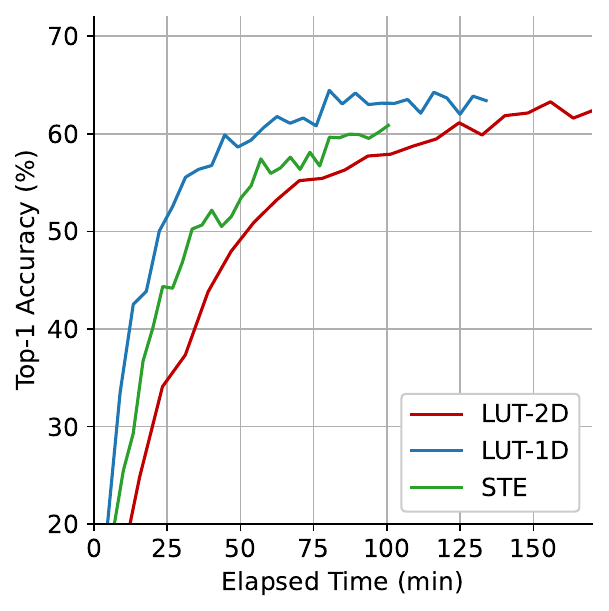}%
}\hspace{0.02\columnwidth}%
\subfigure[Testing accuracy of ResNet50 versus runtime.]{\label{fig:resnet50_time}%
\includegraphics[width=0.48\columnwidth]{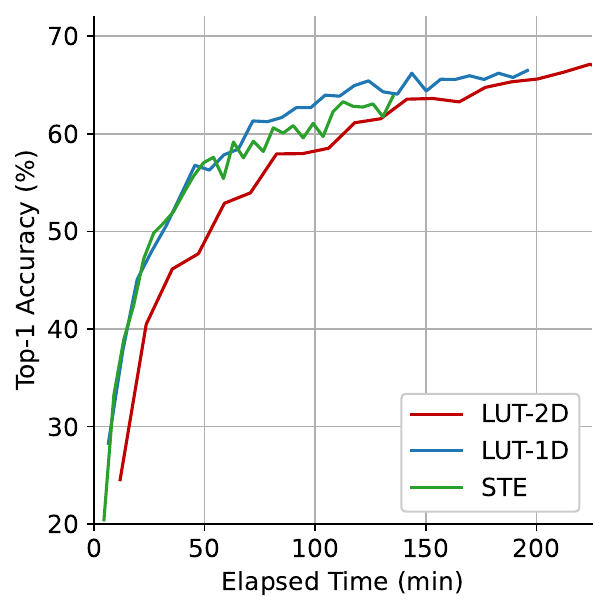}%
}

\caption{Top-1 testing accuracies of ResNet34 and ResNet50 models versus epochs and runtime,
using the 6-bit AppMult mul6u\_rm4 on the CIFAR-100 dataset.}\label{fig:cifar100}
\end{figure}

This experiment compares our gradient approximation method with the STE method on the CIFAR-100 dataset using the ResNet34 and ResNet50 models.
The learning rate is fixed at $5\times 10^{-4}$,
while the other training settings are the same as those in Section~\ref{subsect:res-cifar10}.
To show the wide applicability of our framework,
instead of using 8-bit and 7-bit AppMults,
we test the 6-bit unsigned AppMult in Table~\ref{tab:appmult} (mult6u\_rm4),
which removes the rightmost 4 columns of partial products.
Compared to the 6-bit AccMult,
mul6u\_rm4 saves 27\% area, 17\% delay, and 33\% power consumption.

Fig.~\ref{fig:cifar100} illustrates the top-1 testing accuracy curves for two models, ResNet34 and ResNet50, trained on the CIFAR-100 dataset over 30 epochs.
From Fig.~\ref{fig:resnet34} and Fig.~\ref{fig:resnet50},
in terms of testing accuracy versus epochs, 
across both ResNet34 and ResNet50 models, 
the proposed LUT-1D and LUT-2D methods consistently achieve higher top-1 accuracy than the STE method for all epochs.
The final accuracy of LUT-2D is 63.47\% and 67.89\% for ResNet34 and ResNet50, respectively, 
outperforming LUT-1D slightly and STE by a large margin.
This indicates that LUT-2D provides more precise gradients for approximate multipliers, 
enabling high-accuracy model retraining.

When comparing accuracy versus runtime, 
although STE converges faster in the early phase, 
its final accuracy lags behind.
LUT-1D achieves a good balance between runtime efficiency and final accuracy, 
while LUT-2D, despite requiring slightly longer training, achieves the highest accuracy overall. 
Using the LUT-1D method, the retraining of both ResNet34 and ResNet50 models converges fastest within the same runtime budget,
while the LUT-2D method converges most slowly under the same runtime due to its higher computational overhead.
However, if we allow enough runtime budget,
the LUT-2D method achieves the highest final accuracy.
Based on these observations,
in the following experiments with large deep learning models,
we only test the LUT-1D method due to its better balance between accuracy and runtime,
comparing it with the STE method.

\begin{table*}[!htbp]
\centering
\caption{Retraining results of vision transformer models using the TransAxx framework and our LUT-1D method. 
Power consumption and delay are normalized to the 8-bit signed AccMult. 
{\color[HTML]{00B050} \tbf{Green bold}} entries denote that the LUT-1D method outperforms TransAxx.}
\label{tab:res-vit}
\begin{tabular}{cl|rrr|rr|rrr}
\bottomrule
\multirow{2}{*}{Model} & \multicolumn{1}{c|}{\multirow{2}{*}{Multiplier}} & \multicolumn{3}{c|}{Top-1 accuracy after retraining}     & \multicolumn{2}{c|}{Runtime/hour}   & \multicolumn{3}{c}{Multiplier information} \\
                       & \multicolumn{1}{c|}{}                            & \makecell{TransAxx\\(STE grad.)} & LUT-1D           & \makecell{Improve-\\ment}      & \makecell{TransAxx\\(STE grad.)} & LUT-1D         & \makecell{Norm.\\power} & \makecell{Norm.\\delay}  & NMED/\%   \\ \hline\hline
ViT-S                  & mul8s\_1KV9                                     & 64.49\%  & {\color[HTML]{00B050} \textbf{75.17}\%} & {\color[HTML]{00B050}\textbf{10.68}\%} & 1.26     & {\color[HTML]{00B050} \textbf{1.17}} & 0.92           & 1.08          & 0.0065    \\
(FP32 acc.             & mul8s\_1L2H                                     & 61.58\%  & {\color[HTML]{00B050} \textbf{75.53}\%} & {\color[HTML]{00B050}\textbf{13.95}\%} & 1.26     & {\color[HTML]{00B050} \textbf{1.16}} & 0.67           & 0.86          & 0.081     \\
74.63\%)               & mul8s\_1L2L                                     & 0.48 \%  & {\color[HTML]{00B050} \textbf{69.36}\%} & {\color[HTML]{00B050}\textbf{68.88}\%} & 1.26     & {\color[HTML]{00B050} \textbf{1.16}} & 0.45           & 0.75          & 0.23      \\ \hline
ViT-B                  & mul8s\_1KV9                                     & 52.24\%  & {\color[HTML]{00B050} \textbf{69.65}\%} & {\color[HTML]{00B050}\textbf{17.41}\%} & 4.64     & {\color[HTML]{00B050} \textbf{3.03}} & 0.92           & 1.08          & 0.0065    \\
(FP32 acc.             & mul8s\_1L2H                                     & 46.17\%  & {\color[HTML]{00B050} \textbf{75.55}\%} & {\color[HTML]{00B050}\textbf{29.37}\%} & 4.64     & {\color[HTML]{00B050} \textbf{3.02}} & 0.67           & 0.86          & 0.081     \\
81.10\%)               & mul8s\_1L2L                                     & 0.15 \%  & {\color[HTML]{00B050} \textbf{71.65}\%} & {\color[HTML]{00B050}\textbf{71.50}\%} & 4.64     & {\color[HTML]{00B050} \textbf{3.02}} & 0.45           & 0.75          & 0.23      \\ \hline
DeiT-S                 & mul8s\_1KV9                                     & 72.23\%  & {\color[HTML]{00B050} \textbf{76.72}\%} & {\color[HTML]{00B050}\textbf{4.49} \%} & 1.32     & {\color[HTML]{00B050} \textbf{1.14}} & 0.92           & 1.08          & 0.0065    \\
(FP32 acc.             & mul8s\_1L2H                                     & 70.72\%  & {\color[HTML]{00B050} \textbf{77.19}\%} & {\color[HTML]{00B050}\textbf{6.47} \%} & 1.32     & {\color[HTML]{00B050} \textbf{1.18}} & 0.67           & 0.86          & 0.081     \\
81.04\%)               & mul8s\_1L2L                                     & 62.15\%  & {\color[HTML]{00B050} \textbf{74.83}\%} & {\color[HTML]{00B050}\textbf{12.67}\%} & 1.32     & {\color[HTML]{00B050} \textbf{1.12}} & 0.45           & 0.75          & 0.23      \\ \hline
Swin-S                 & mul8s\_1KV9                                     & 68.14\%  & {\color[HTML]{00B050} \textbf{80.24}\%} & {\color[HTML]{00B050}\textbf{12.09}\%} & 3.15     & 3.46          & 0.92           & 1.08          & 0.0065    \\
(FP32 acc.             & mul8s\_1L2H                                     & 66.88\%  & {\color[HTML]{00B050} \textbf{80.92}\%} & {\color[HTML]{00B050}\textbf{14.03}\%} & 3.15     & 3.43          & 0.67           & 0.86          & 0.081     \\
83.06\%)               & mul8s\_1L2L                                     & 56.21\%  & {\color[HTML]{00B050} \textbf{78.89}\%} & {\color[HTML]{00B050}\textbf{22.68}\%} & 3.15     & 3.52          & 0.45           & 0.75          & 0.23      \\ \hline \hline
\multicolumn{2}{c|}{\makecell{Mean over all models\\\& all AppMults}}    & 51.79\%  & {\color[HTML]{00B050} \textbf{75.47}\%} & {\color[HTML]{00B050}\textbf{23.69}\%} & 2.59     & {\color[HTML]{00B050} \textbf{2.20}} &                &               &           \\ \bottomrule
\end{tabular}
\end{table*}

\subsection{Experiments on AppMult-Based Vision Transformers and ImageNet Dataset}\label{subsect:res-vit}
This set of experiments evaluates the effectiveness of our proposed AppMult gradient estimation methods on vision transformer models.
All runs in this set of experiments use a single NVIDIA A100 GPU.
The baseline method is the state-of-the-art one, TransAxx~\cite{danopoulos2025transaxx},
which is a retraining framework for vision transformers using AppMults.
TransAxx uses the STE gradient estimation method for AppMults,
which estimates the gradient of the AppMult using that of the AccMult.
The tested models are ViT-S~\cite{dosovitskiy2020image}, ViT-B~\cite{dosovitskiy2020image}, Deit-S~\cite{touvron2021training}, and Swin-S~\cite{liu2021swin}.
For the above models,
our framework applies 8-bit symmetric layer-wise quantization to both the model weights and the activations in all the linear and convolutional layers.
After quantization,
all the 8-bit AccMults are replaced with 8-bit signed AppMults.
The tested AppMults are three 8-bit signed AppMults from EvoApproxLib~\cite{mrazek2017evoapprox8b},
\ie{}, mul8s\_1KV9, mul8s\_1L2H, and mul8s\_1L2L,
which are the same as those used in TransAxx.
The applied dataset is the ImageNet~\cite{deng2009imagenet}.

For efficient retraining,
we apply a similar strategy as in TransAxx
and only select a subset of 100,000 images from the training set for retraining.
We re-run the open-source code of TransAxx~\cite{danopoulos2025transaxxcode} and compare the results with our LUT-1D method.
For efficiency, we do not run our LUT-2D method on the vision transformer models.
For both TransAxx and our LUT-1D method,
the retraining parameters are as follows:
the learning rate is fixed as $5\times 10^{-5}$,
the batch size is 128,
the retraining epochs is 5,
and the optimizer is Adam.
We apply layer-wise asymmetric quantization using Eq.~\eqref{eq:observer} for both weights and activations in all the linear and convolutional layers.

Table~\ref{tab:res-vit} reports the top-1 accuracy after retraining, total retraining runtime in hours for each model and each AppMult.
Multiplier characteristics are also reported,
including normalized power consumption and delay (both normalized to the 8-bit signed AccMult) and NMED.
Bold entries denote that our proposed LUT-1D method outperforms TransAxx.
The multiplier information columns summarize hardware-error trade-offs. 
The mul8s\_1KV9 AppMult is near-accurate (normalized power 0.92, normalized delay 1.08, NMED 0.0064\%). 
The mul8s\_1L2H AppMult provides larger savings (normalized power 0.67, normalized delay 0.86) with higher error (NMED 0.081\%). 
The mul8s\_1L2L AppMult is the most aggressive (normalized power 0.45, normalized delay 0.75) and has the highest error (NMED 0.23\%).

Across all models and multipliers, LUT-1D consistently yields higher accuracy than TransAxx. 
With better gradient estimation of the AppMults through the LUT-1D method,
our LUT-1D method
improves the retraining accuracy by 23.69\% on average across all models and multipliers,
compared to TransAxx.
The smallest improvement is 4.49\% for the DeiT-S model with the mul8s\_1KV9 AppMult,
while the largest improvement is 71.50\% for the ViT-B model with the mul8s\_1L2L AppMult.
For ViT-S, LUT-1D even slightly exceeds the floating-point 32-bit reference accuracy (74.63\%) when using the mul8s\_1KV9 and mul8s\_1L2H AppMults,
and the power consumption of the multiplier is reduced by 8\% and 33\%, respectively.
Notably, for the large-error multiplier mul8s\_1L2L,
on the ViT-S and ViT-B models,
TransAxx fails to recover the accuracy through retraining,
while our LUT-1D method achieves significant improvements of 68.88\% and 71.50\% respectively, compared to TransAxx.
This demonstrates the effectiveness of our LUT-1D method in recovering accuracy for large-error multipliers.

Regarding the runtime,
our LUT-1D method is more efficient than TransAxx for the ViT-S, ViT-B, and DeiT-S models,
while LUT-1D is slightly slower for the Swin-S model.
However, for the Swin-S model,
the runtime of LUT-1D is still acceptable,
and the accuracy improvement is significant.
On average, our LUT-1D method is 1.2x faster than TransAxx across all models and multipliers.



\subsection{Exploration on Extremely Low-Bit AppMults}\label{subsect:res-extreme}

This set of experiments explores the effectiveness of our proposed gradient estimation methods under extremely low-bit quantization.
We test our framework on the ImageNet dataset using the ResNet18 and ResNet50 models.
As in Section~\ref{subsect:res-nn},
we replace all the AccMults in the convolutional layers with the same type of AppMults.

The training parameters are as follows:
batch size 128, number of epochs 10, Adam optimizer, and initial learning rate $5\times 10^{-5}$.
A cosine annealing learning rate scheduler is applied to adjust the learning rate during retraining.
We apply 4-bit asymmetric channel-wise quantization to both the weights and the activations in all the linear and convolutional layers.
The quantization method is the learnable clipping method in~\cite{shao2023omniquant}.
The applied AppMult is a 4-bit unsigned AppMult, \ie{}, mul4u\_rm2 that removes the 2 LSB columns of the exact 4-bit unsigned multiplication.
Compared to the 4-bit AccMult, mul4u\_rm2 saves 17.3\% area, 17.8\% delay, and 20.8\% power.
The whole ImageNet training set is used for retraining to improve the model accuracy.

For efficiency, we only test the LUT-1D method in this set of experiments.
Table~\ref{tab:low-bit} presents the retraining results of ResNet18 and ResNet50 on ImageNet using the 4-bit unsigned AppMult mul4u\_rm2. 
Before retraining, both networks suffer from almost complete accuracy loss (0.10\%), indicating the severe impact of low-bit approximation. After retraining, the accuracy is largely recovered, 
and our proposed LUT-1D method consistently outperforms the baseline STE approach. Specifically, LUT-1D achieves a 0.29\% improvement for ResNet18 and a 0.35\% improvement for ResNet50, 
demonstrating its effectiveness in improving model robustness under aggressive approximation.

\begin{table}[!htbp]
\centering
\caption{Results on 4-bit unsigned AppMult (\tit{mul4u\_rm2}) with ResNet18 and ResNet50 on ImageNet.
The {\color[HTML]{00B050} \textbf{bold}} entries denote that our proposed LUT-1D method outperforms the STE method.}
\tabcolsep=4pt
\label{tab:low-bit}
\begin{tabular}{cccccc}
\toprule
         &            & \multicolumn{3}{c}{Acc. after retrain} \\
\multirow{-2}{*}{CNN} & \multirow{-2}{*}{\makecell{Acc. before\\retrain}} & STE & LUT-1D & Impr. \\\midrule
\makecell{ResNet18\\(fp32 acc. 71.00\%)} & 0.10\% & 67.30\% & {\color[HTML]{00B050} \textbf{67.59\%}} & {\color[HTML]{00B050} \textbf{0.29\%}} \\\hline
\makecell{ResNet50\\(fp32 acc. 76.62\%)} & 0.10\% & 74.03\% & {\color[HTML]{00B050} \textbf{74.38\%}} & {\color[HTML]{00B050} \textbf{0.35\%}} \\ \bottomrule
\end{tabular}
\end{table}
\section{Conclusion}\label{sect:concl}

We presented two gradient estimation methods for AppMults to improve retraining accuracy of deep learning models. 
The LUT-2D method provides fine-grained gradients and achieves the highest accuracy. 
The LUT-1D method is a compact variant that reduces memory and lookup runtime, while still achieving high accuracy.
Across CNNs and vision transformers,
both methods improve retraining accuracy compared to the state-of-the-art STE method.

\section*{Acknowledgement}

\cadd{This work is supported in part by Synopsys Inc..}

\printbibliography

\begin{IEEEbiography}[{\includegraphics[width=0.90in,height=1.10in,clip,keepaspectratio]{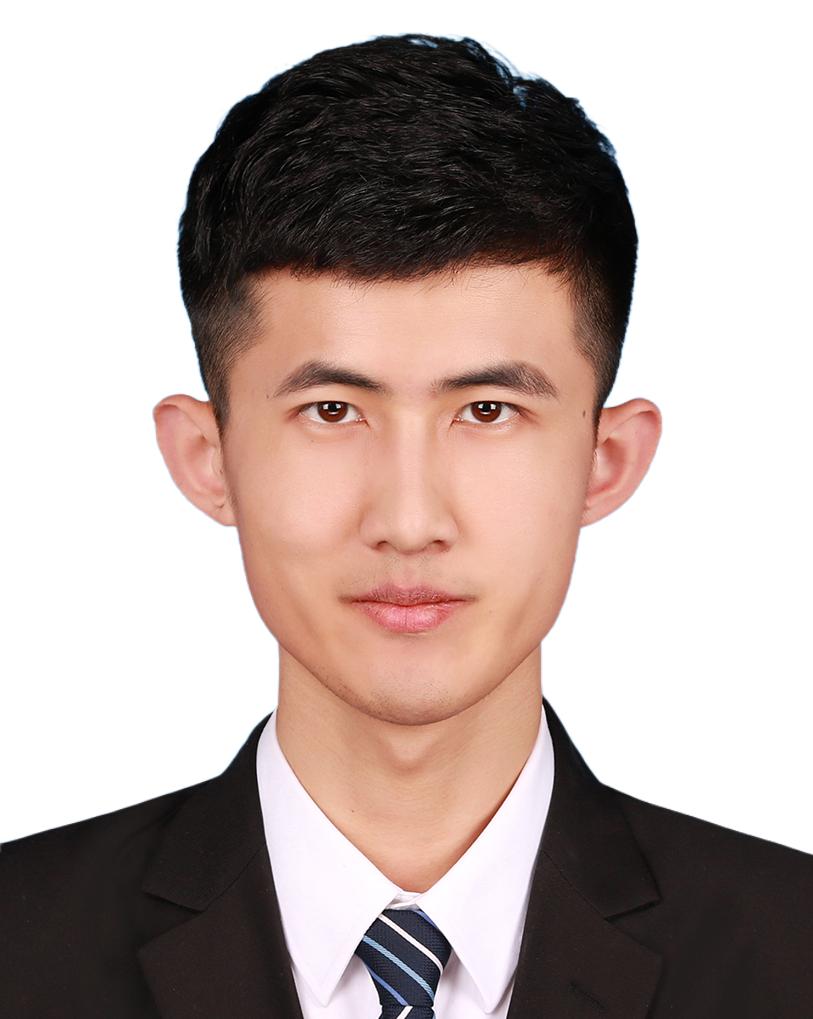}}]{Chang Meng}
    is an Assistant Professor in the Department of Mathematics and Computer Science at Eindhoven University of Technology, Eindhoven, the Netherlands.
    He was a postdoctoral researcher at the Integrated Systems Laboratory, EPFL Lausanne, Switzerland, from 2023 to 2025.
    He received his Ph.D. degree in Electronic Science and Technology at Shanghai Jiao Tong University in 2023.
    His research interest is electronic design automation for emerging computing paradigms,
    especially the logic synthesis and verification of approximate computing circuits.
    His research works were nominated for the Best Paper Award at Design, Automation, and Test in Europe Conference (DATE),
    and received the Best Paper Award at the International Workshop on Logic Synthesis (IWLS) and the International Conference on Synthesis, Modeling, Analysis and Simulation Methods, and Applications to Circuit Design (SMACD).
\end{IEEEbiography}

\begin{IEEEbiography}[{\includegraphics[width=0.90in,height=1.10in,clip,keepaspectratio]{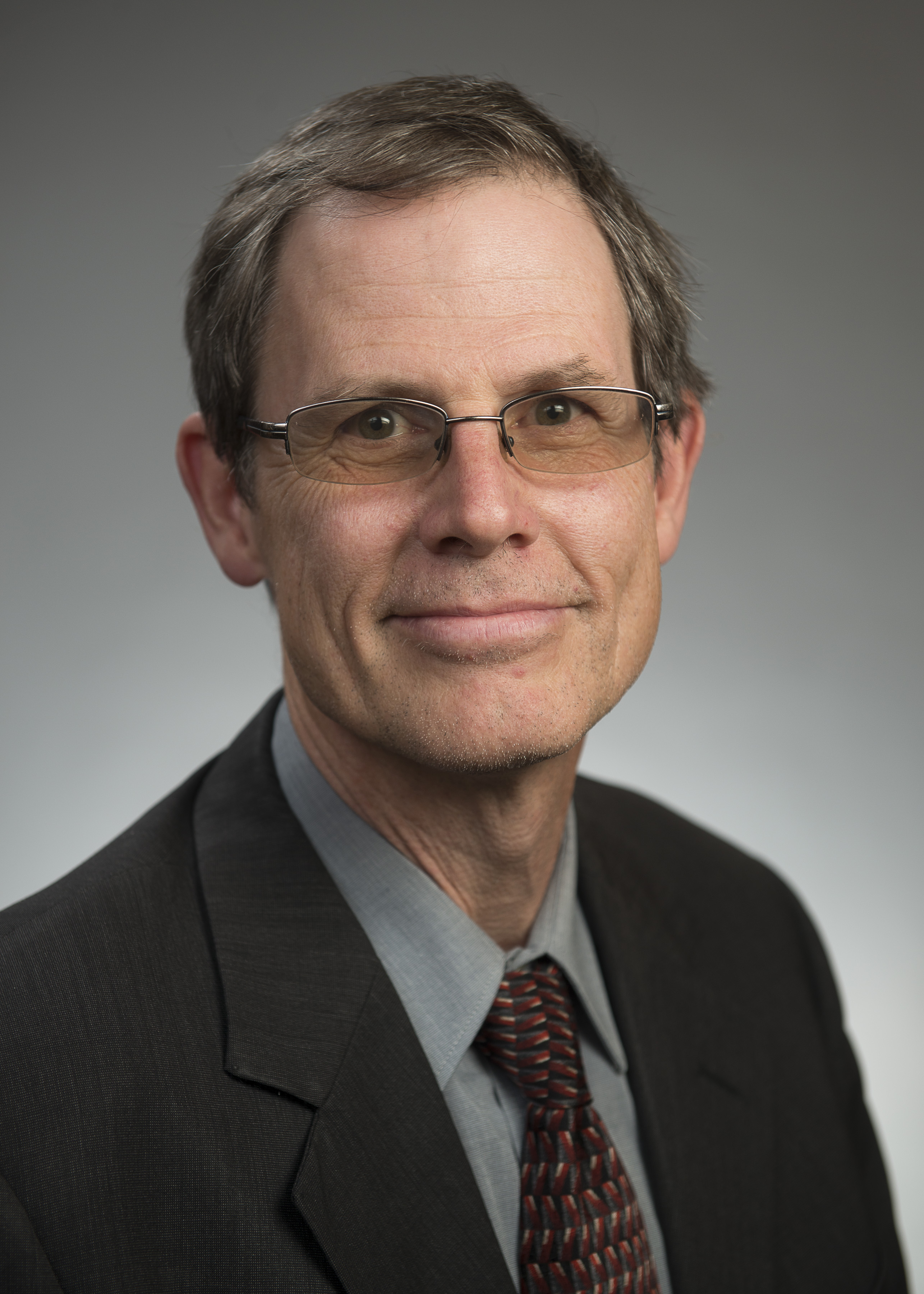}}]{Wayne Burleson}
    \cadd{is Professor and Graduate Program Director of Electrical and Computer Engineering at the University of Massachusetts Amherst, MA, USA, where he has been a faculty member since 1990.
    He received  B.S. and M.S. degrees in E.E. from the Massachusetts Institute of Technology, Cambridge, MA, USA, and the Ph.D. degree in E.C.E. from the University of Colorado, Boulder, CO, USA.

    He currently directs and conducts research in VLSI and security engineering with funding from NSF and ARL. He worked for AMD Research from 2012-2017 on Exascale Computing. 
    He is a Fellow of the IEEE and recipient of the 2024 ACM/IEEE CEDA A. Richard Newton Technical Impact Award in EDA. 
    He was also Distinguished Lecturer and recipient of the University of Massachusetts Amherst Chancellor's Award.
    Although his primary focus is now on hardware security and power-efficient acceleration, building on 40 years in the microelectronics field, 
    he also studies higher level issues of system security and security economics, with applications in payment systems, radio-frequency identification and medical devices. 
    He teaches courses in microelectronics, embedded systems, and security engineering.}
\end{IEEEbiography}

\begin{IEEEbiography}[{\includegraphics[width=0.90in,height=1.10in,clip,keepaspectratio]{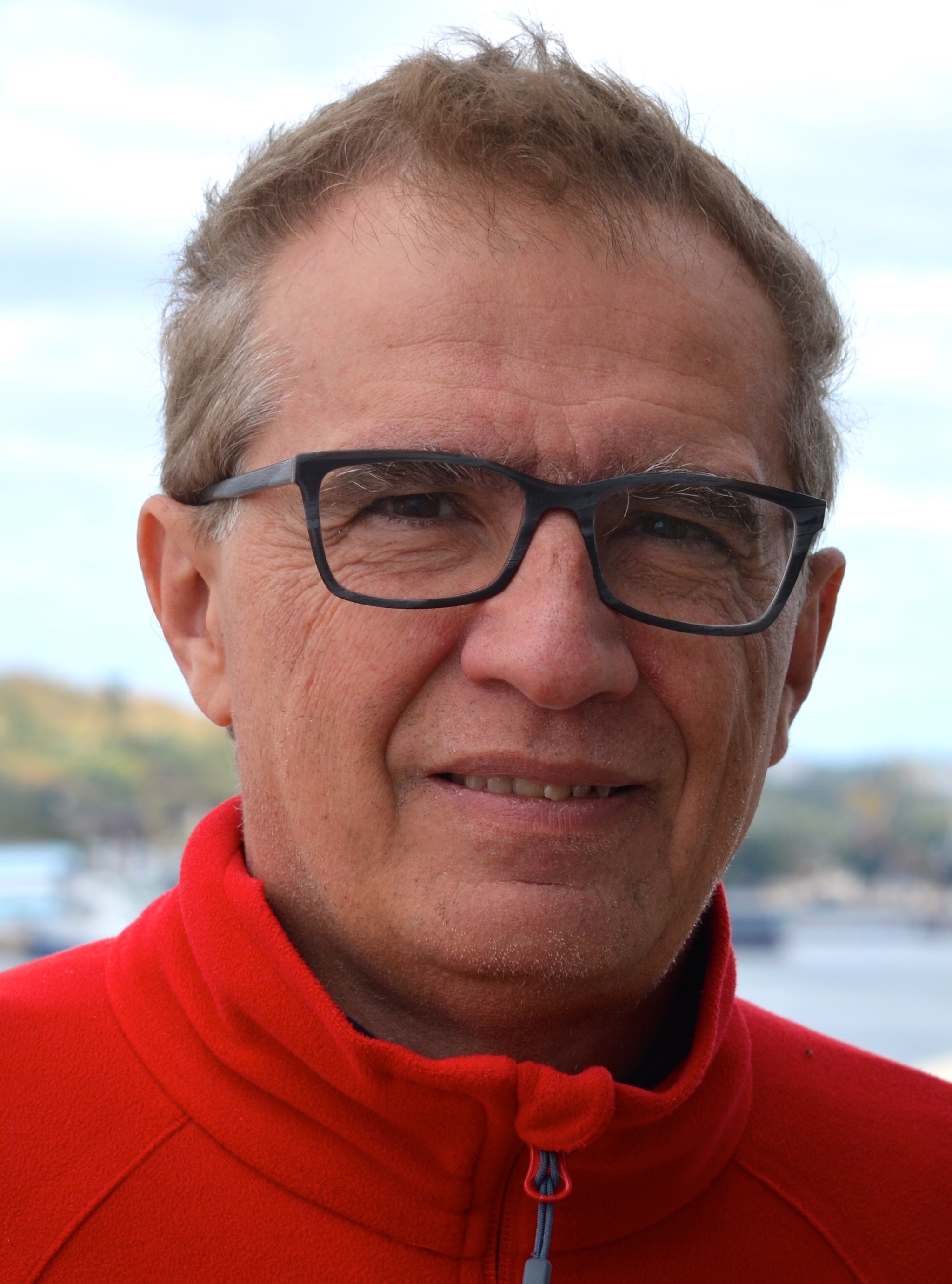}}]{Giovanni De Micheli}
    is Professor and Director of the Integrated Systems Laboratory and Scientific Director of the EcoCloud center at EPFL Lausanne, Switzerland.
    Previously, he was Professor of Electrical Engineering at Stanford University.
    He holds a Nuclear Engineer degree (Politecnico di Milano, 1979), 
    a M.S. and a Ph.D. degree in Electrical Engineering and Computer Science (University of California at Berkeley, 1980 and 1983).

    He is a Fellow of ACM, AAAS and IEEE, 
    a member of the Academia Europaea, of the Swiss Academy of Engineering Sciences, and International Honorary member of the American Academy of Arts and Sciences.
    His current research interests include several aspects of design technologies for integrated circuits and systems, such as synthesis for emerging technologies. 
    He is also interested in heterogeneous platform design including electrical components and biosensors, as well as in data processing of biomedical information.
    He is member of the Scientific Advisory Board of IMEC (Leuven, B) and STMicroelectronics.

    Professor De Micheli is the recipient of the 2025 IEEE Gustav Kirchhoff Award,
    the 2022 ESDA-IEEE/CEDA Phil Kaufman Award, 
    the 2019 ACM/SIGDA Pioneering Achievement Award, and several other awards.
\end{IEEEbiography}

\end{document}